%% file: arxiv.tex
\documentclass[10pt,twocolumn,letterpaper]{article}

\usepackage[accsupp]{axessibility}  
\usepackage[pagenumbers]{cvpr} 
\usepackage{fancyhdr}  
\input{preamble}

\definecolor{cvprblue}{rgb}{0.21,0.49,0.74}
\usepackage[pagebackref,breaklinks,colorlinks,allcolors=cvprblue]{hyperref}

\def\paperID{4} 
\def\confName{XAI4CV}
\def\confYear{2026}
\title{FAME: Feature Activation Map Explanation on Image Classification and Face Recognition}

\author{Xinyi Zhang \qquad Manuel G\"unther\\
Department of Informatics, University of Zurich\\
{\tt\small xinyi.zhang@uzh.ch, siebenkopf@googlemail.com}
}
\begin{document}
\maketitle
\fancyhf{} 
\fancyhead[C]{\small This is a pre-print of the original paper accepted by the Conference on CVPR Workshop (CVPRW) 2026.}
\thispagestyle{fancy}

\input{sections/abstract}
\input{sections/introduction}
\input{sections/related_work}
\input{sections/approach}
\input{sections/experiments}
\input{sections/conclusion.tex}
\section*{Acknowledgements}
This work is supported by the University of Zurich via the UZH Candoc Grant, grant no. FK-25-016.

{
    \small
    \bibliographystyle{ieeenat_fullname}
\bibliography{sections/References,sections/Publications}
}

\clearpage
\appendix
\twocolumn[{%
  \vspace{0.5em}
  \begin{center}
    {\Large\textbf{Supplemental Material}}
  \end{center}
  \vspace{0.5em}
}]
\input{sections/supplemental}

\end{document}

%% file: preamble.tex
\usepackage{framed,multirow,tabularx}

\newcommand{\figref}[1]{Fig.~\ref{#1}\xspace}
\newcommand{\subfigref}[1]{Fig.~\subref{#1}\xspace}
\newcommand{\tabref}[1]{Tab.~\ref{#1}\xspace}
\newcommand{\subtabref}[1]{Tab.~\subref{#1}\xspace}
\newcommand{\secref}[1]{Sec.~\ref{#1}\xspace}

\usepackage{amsmath}

\newcommand\Caption[3][]{\caption[#2]{\label{#1}\textsc{#2}.\small#3}}
\newcommand{\rednum}[1]{\textbf{#1}}



%% file: sections/abstract.tex
\begin{abstract}
Deep Learning has revolutionized machine learning, reaching unprecedented levels of accuracy, but at the cost of reduced interpretability.
Especially in image processing systems, deep networks transform local pixel information into more global concepts in a highly obscured manner.
Explainable AI methods for image processing try to shed light on this issue by highlighting the regions of the image that are important for the prediction task.
Among these, Class Activation Mapping (CAM) and its gradient-based variants compute attributions based on the feature map and upscale them to the image resolution, assuming that feature map locations are influenced only by underlying regions.
Perturbation-based methods, such as CorrRISE, on the other hand, try to provide pixel-level attributions by perturbing the input with fixed patches and checking how the output of the network changes.
In this work, we propose Feature Activation Map Explanation (FAME), which combines both worlds by using network gradients to compute changes to the input image, manipulating it in a gradient-driven way rather than using fixed patches.
We apply this technique on two common tasks, image classification and face recognition, and show that CAM's above-mentioned assumption does not hold for deeper networks.
We qualitatively and quantitively show that FAME produces attribution maps that are competitive state-of-the-art systems.
Our code is available: {\footnotesize \url{https://github.com/AIML-IfI/fame}.}
\end{abstract}

%% file: sections/introduction.tex
\section{Introduction}
\label{sec:intro}

Image processing networks extract information from an image in highly non-linear and obscured ways to reach their excellent performance in several tasks, such as Image Classification (IC) and Face Recognition (FR).
While early networks \cite{krizhevsky2012alexnet} were shallow enough for partial layer-wise visualization \cite{zeiler2014visualizing}, modern architectures with hundreds of layers are far too complex to inspect individually.
Therefore, research on eXplainable AI (XAI) has moved toward attribution methods, only highlighting the regions in the image that are influential to the decision process.

There exist two main streams of attribution methods.
Gradient-based Class Activation Mapping (Grad-CAM) \cite{selvaraju2017gradcam} and its variants make use of gradient information (see \secref{sec:grad-cam}) and feature map activations to produce attribution maps.
Since attributions are created in the size of the feature map, they are upscaled to image resolutions to provide pixel-level attributions.
This upscaling is based on the assumption that elements in the feature map are influenced only by the underlying pixels.
However, receptive fields of deep networks are much larger, and we show in our experiments that this assumption does not hold.
Perturbation-based methods \cite{ivanovs2021perturbation}, on the other hand, do not make use of the network's gradient.
Instead, they perturb local parts of the inputs and test how the prediction changes.
For example, in an image classification task, when important parts of the input are perturbed, the prediction of the correct class should reduce \cite{fong2017interpretable}.
However, the correct way to perturb images is unclear.
For example, when adding local black squares or noise, such perturbations induce sharp edges into the input, which might represent features of some classes.
When blurring local regions, this does not necessarily remove important information as modern networks are well-able to handle highly-blurred images \cite{robbins2024daliid}.

In our proposed Feature Activation Map Explanation (FAME) we combine the two aspects by using network gradients that go back to the input level \cite{huber2024xssab,lu2024fggb} to compute attribution maps at the pixel level.
Particularly, we exploit an iterative method that we had originally developed as adversarial attack \cite{rozsa2017lots}, where we modify the original image in a more principled way in order to change network outputs, and interpret adversarial perturbations as an attribution map.

Depending on the target layer where we extract the network's output, and on the loss function used to obtain gradients, FAME can be applied to highlight various particularities.
We start by extracting elements from the feature map and show that their receptive field is much larger for deeper networks than anticipated by CAM-based methods.
Successively, we apply FAME on the predictions of classification networks, producing more fine-grained attribution maps that capture model-relevant sensitivity patterns beyond CAM-based methods.
Finally, we show the generality of our method by highlighting similar and dissimilar regions of a pair of faces involved in a verification process.
We quantitatively evaluate FAME against state-of-the-art methods and show that it is able to highlight important regions in the image similarly to or better than competitors.
We also qualitatively show that FAME attributions cover more and better regions of objects than related methods.

%% file: sections/related_work.tex
\section{Related Work}
\label{sec:related}

\subsection{Deep Learning and the Black-Box Problem}
Deep neural networks have achieved state-of-the-art performance across vision tasks such as Image Classifications (IC) \cite{he2016deep} and Face Recognition (FR) \cite{deng2019arcface}.
In IC, networks $f$ learn to hierarchically extract discriminative visual features $\varphi$ and predict the logits $z$ that assign semantic labels to entire images.
In FR, given a pair of images, one from gallery $x_p$, and one from probe $x_p$, FR systems $f$ learn to embed facial representations in a compact feature space, and compute the similarity to distinguish if they are from the same identity \cite{phillips2011evaluation}.
However, these deep networks are often considered as black boxes, making it difficult to understand the reasoning behind their predictions or detect potential biases and vulnerabilities \citep{tucci2024overview}.
This lack of transparency complicates the understanding of the performance degradations of IC and FR systems, which decrease under extreme conditions, such as variations in pose, quality, and occlusion.

\subsection{Explainable AI techniques}
EXplainable Artificial Intelligence (XAI) aims to make deep learning models more interpretable by addressing their black-box nature \cite{nauta2023anecdotal,longo2024xai2.0}.
Generally, XAI approaches are divided into intrinsic methods that design inherently interpretable models \cite{rudin2018stop}, and post-hoc methods that explain the behavior of complex pre-trained models without modifying their structure \cite{li2025diffcam}.
The latter have become the dominant choice for explaining deep neural networks due to their flexibility and scalability.
Amongst these, Class Activation Mapping (CAM) \cite{selvaraju2017gradcam} visualizes discriminative regions by backpropagating gradients from the output to feature maps.
Its extensions, such as Grad-CAM++ \cite{chattopadhay2018grad-cam++} and HiResCAM \cite{draelos2020hirescam}, improve spatial precision.
FullGradCAM \cite{srinivas2019fullgrad} aggregates full-gradient contributions from both the input and all intermediate feature maps, thereby incorporating multi-layer information beyond single-layer approaches like Grad-CAM.
Lately, research has turned to make use of gradients to the input \cite{lu2024fggb,huber2024xssab}, see \secref{sec:fggb}.

In contrast, perturbation-based methods generate explanations by measuring the impact of input modifications on model outputs.
For example, \citet{lu2024corrrise} proposed Corr\-RISE to explain deep face verification models.
They extended the RISE framework \cite{petsiuk2018rise} to the face verification setting by computing Pearson correlation between randomly masked input face images with the similarity scores between two images.
CorrRISE produces two saliency maps that highlight regions contributing to higher (\emph{similar}) or lower (\emph{dissimilar}) similarity.
Although CorrRISE provides a baseline for explaining similarity decisions, it relies on random input masking, which introduces stochasticity and depends on manually-defined mask parameters.

\subsection{Evaluation Metrics for XAI Techniques}
Unfortunately, there is no universally-accepted metric for evaluating visual explanations, and different tasks often require different evaluation criteria.
\citet{adebayo2018sanity} even question whether using one AI model to validate another provides principled guarantees, showing that certain saliency maps are insensitive to model parameters.
For IC, Intersection over Union (IoU) \cite{zhou2016localization} is commonly used to measure the spatial overlap of attribution heatmaps with ground-truth regions, suitable for localized explanations.
Remove and Debias (ROAD) \cite{rong2022road} mitigates confounding effects in removal-based evaluations, improving reliability without requiring model retraining.
Deletion and Insertion curves \cite{petsiuk2018rise} evaluate how model confidence changes as salient pixels are progressively removed from the original image, or added to an empty canvas.
\citet{lu2024corrrise} extended the Delete and Insert metric for face verification.

%% file: sections/approach.tex
\section{Method}
\label{sec:method}

Our explainable AI method, the Feature Activation Map Explanation (FAME) is built on the basis of our Layerwise Origin Target Synthesis (LOTS) \cite{rozsa2017lots} that we originally designed for adversarial attacks.
LOTS iteratively computes pixel-level perturbations to the original image such that the network output at a certain layer represents a specific target.
We reinterpret this method and turn it into an explainable AI technique, providing fine or coarse input-level attributions.

Subsequently, we make use of the following notation.
For a given input image $x\in\mathbb R^{C\times H\times W}$ with $C$ color channels and spatial dimensions $H\times W$, a deep network $f(x)$ is composed of $l=1,\ldots,L$ layers $f^l(\cdot)$ which are concatenated to produce $O$ output logits $z\in\mathbb R^O$:
\begin{equation}
  \label{eq:network}
  z = f^L\biggl(f^{L-1}\Bigl(f^{L-2}\bigl(\cdots f^{1}(x)\cdots\bigr)\Bigr)\biggr)
\end{equation}
We particularly look into the output of two layers.
One is the last convolutional layer $a=f^{l_a}(\cdot)$,\footnote{Similar concepts exist for Vision Transformer networks, but we leave the evaluation of these to future work.} which is typically $l_a = L-2$, that extracts the feature map $a \in \mathbb R^{C_a\times H_a\times W_a}$ with increased channels $C_a\gg C$ and reduced spatial dimensionality $H_a\times W_a\ll H\times W$.
The other layer of interest is the embedding layer $\varphi = f^{l_\varphi}(\hat a)$ (typically $l_\varphi=L-1$) that processes a transformed feature map $\hat a$ into an embedding $\varphi \in \mathbb R^{C_\varphi}$.
For IC, the feature map transform is often a global pooling with $\hat a \in \mathbb R^{C_a}$ and the logits $z$ are typically used to predict the class of interest.
For FR, the feature map $a$ is flattened to $\hat a\in \mathbb R^{C_a\cdot H_a\cdot W_a}$ and specialized logits \cite{deng2022arcface,meng2021magface,kim2022adaface} are used to train discriminative features $\varphi$.
Features $\varphi_g$ and $\varphi_p$ are extracted for a gallery $x_g$ and a probe image $x_p$ and compared using cosine, \ie, the dot product of the normalized embeddings:\\[-1ex]
\begin{equation}
  \label{eq:cos}
  s = \frac{\varphi_g^\top \varphi_p}{\|\varphi_g\|\,\|\varphi_p\|} = \left(\frac{\varphi_g}{\|\varphi_g\|}\right)^\top\!\left(\frac{\varphi_p} {\,\|\varphi_p\|}\right).
\end{equation}


\subsection{Grad-CAM}
\label{sec:grad-cam}
Gradient-based Class Activation Mapping (Grad-CAM) makes use of the gradient \cite{selvaraju2017gradcam}.
In the case of image classification, this gradient is the partial derivative $\partial z_o / \partial a[k]$ of the logit for a given class $o\in[1,O]$ to a specific element $a[k] \in \mathbb R^{C_a}$ of the feature map with the location index $k\in[1,H_a]\times[1,W_a]$.
These derivatives are combined with the feature map in various ways \cite{chattopadhay2018grad-cam++,draelos2020hirescam} to produce an intermediate attribution map $\check e \in \mathbb R^{H_a\cdot W_a}$, which is then max-normalized and bilinearly interpolated to $e \in \mathbb R^{H\cdot W}$.

Grad-CAM was also extended to explain FR by backpropagating the similarity score $s$.
Specifically, to avoid including the normalization factor from \eqref{eq:cos}, \citet{zhu2021visual} compute $\partial \varphi_g^\top \varphi_p / \partial a_p[k]$ and $\partial \varphi_g^\top \varphi_p / \partial a_g[k]$.

\subsection{FGGB}
\label{sec:fggb}
Recently, research has turned to using backpropagation of face embeddings $\varphi$ to the input to compute attribution maps $e$.
For example, \citet{huber2024xssab} compute the Hadamard product between two embeddings: $v = \varphi_g \odot \varphi_p$, which they split into $v_+$ and $v_-$, depending on whether $v_i$ exceeds a specific threshold $\theta$.
They backpropagate the average $v_+$ (and $v_-$) separately: $e_+ = \partial v_+/\partial x_p$.
Based on this, \citet{lu2024fggb} proposed Feature-Guided Gradient Backpropagation (FGGB) by backpropagating each $v_i$ to produce different $e_i$ per embedding dimension $i$ separately, and normalize the attribution maps individually.
They compute weighted averages $e=\sum_i (v_i -\theta) e_i$, which can be split into similar and dissimilar maps $e_+$ and $e_-$ by thresholding $e$ at 0.

Both works make use of a fixed threshold $\theta$ that is selected based on the EER \cite{huber2024xssab}.\footnote{Actually, \citet{lu2024fggb} does not explain how $\theta$ is selected, so we assume that this is identical to \cite{huber2024xssab}.}
While FAME explores a similar idea, it does not rely on such an arbitrary threshold.
Additionally, previous methods \cite{lu2024fggb,huber2024xssab} are constrained to a particular task, face verification, whereas FAME can be adapted to various attribution tasks.

\subsection{LOTS}
\label{sec:LOTS}
The Layerwise Origin Target Synthesis (LOTS) developed by \citet{rozsa2017lots} is used to create an adversarial image $\bar x\in\mathbb R^{C\times H\times W}$ that fools the network to change its prediction toward a specific target $t_l$ at layer $l$ of the network.
Starting from $\bar x=x$, LOTS makes use of the gradient of the loss $\mathcal L$, which is normalized by the maximum absolute value of the gradient, to update $\bar x$ iteratively:
\begin{align}
  \label{eq:lots}
  \bar x &\leftarrow \bar x - \eta \frac{\nabla_{\bar x}}{\max|\nabla_{\bar x}|} & \nabla_{\bar x} &= \frac{\partial\, \mathcal L \bigl(f^l(\bar x)\mid t^l\bigr)}{\partial \bar x},
\end{align}
with a small step size $\eta>0$.
To create a FR attack, the adversarial probe image $\bar x_p$ is modified such that embedding $\bar \varphi_p = f^{l_\varphi}(\bar x_p)$ gets closer to the chosen gallery image $t_\varphi = f^{l_\varphi}(x_g) = \varphi_g$ using Euclidean loss $\mathcal L_2$ \cite{rozsa2017lots}.
Importantly, LOTS uses the normalized raw gradient in \eqref{eq:lots}, providing a different perturbation for each pixel, and not the sign as in the famous Fast Gradient Sign attack \cite{goodfellow2015explaining} and its variations.

\subsection{FAME}
\label{sec:fame}
Taking a closer look at the result of LOTS, the optimization procedure answers the question: \emph{How do the pixels in the input image need to change to obtain the target as output?}
This can be turned into the question: \emph{Which pixels in the input image are most important to change the prediction?} by computing the adversarial perturbation:
\begin{equation}
  \label{eq:fame-diff}
  \Delta x = \bigl|\bar x - x\bigr|,
\end{equation}
and converting it to grayscale.\footnote{Theoretically, we could also compute different attribution maps for the different color channels, but we leave this to future work.}
In order to turn this pixel-level annotation to a more broad region annotation, and in correspondence with \cite{huber2024xssab,lu2024fggb}, the perturbation map $\Delta x$ is then smoothed using a Gaussian kernel (see \secref{sec:blur} for the effect of different blur kernel sizes), and max-normalized to provide an attribution map $e=[0,1]^{H\times W}$.
Through selecting specific target values $t$ and loss functions $\mathcal L$, we apply FAME to produce attribution maps for different purposes:

\noindent\textbf{FAME for Feature Maps.}
One important question that we ask in our evaluation is: \emph{Which pixels from the input image contribute to single elements $a[k]$ in the feature map?}
Such information is required to assess the viability of the assumption of Grad-CAM, \ie, that only pixels under the feature map location influence the feature and, therefore, we can upscale the attribution map $\check e$ defined in \secref{sec:grad-cam}.
As the target $t$ in \eqref{eq:lots}, we select the zero vector.
While this seems unintuitive upfront, we want to know: \emph{Which image pixels have to be removed in order to extract no information at $a[k]$?}
This is semantically similar to: \emph{Which pixels contain information that influence $a[k]$?}

For each feature map location $k$, we penalize only the activation at that spatial location and ignore all others:
\begin{equation}
\label{eq:fame-fm}
\mathcal{L}_{a}(a[k] \mid 0) = \big\| a[k] - 0 \big\|_{1}.
\end{equation}
Applying FAME to $\mathcal{L}_{a}$ in \eqref{eq:lots} yields a local attribution at location $k$.
Repeating this process over all locations $k$ produces spatially resolved feature-attribution maps.

\noindent\textbf{FAME for Image Classification.}
\label{sec:fame-classification}
We follow the idea of Grad-CAM and define our loss as the logit of a given class $o$ directly: $\mathcal L_{\mathrm{cls}} = z_o$, which we backpropagate via \eqref{eq:lots}.
Unlike CAM-based methods that localize attribution regions in intermediate layers, FAME directly highlights input pixel. 

\noindent\textbf{FAME for Face Recognition.}
\label{sec:fame-face}
We make use of two normalized\footnote{We do not compute gradients for the denominators in \eqref{eq:cos}.} embeddings $\varphi_g$ and $\varphi_p$ extracted from gallery $x_g$ and probe image $x_p$, respectively, from which we calculate the similarity $s$ via \eqref{eq:cos}.
Similar to \citet{zhu2021visual}, we directly minimize the similarity itself to obtain the regions $e_+$ that are important in both images.
Following \cite{lu2024fggb,huber2024xssab,lu2024corrrise}, we estimate dissimilar regions $e_-$ by minimizing $\mathcal L_-$, but without requiring any particular threshold:
\begin{align}
\label{eq:fame-fr-l1}
  \mathcal{L}_{+}(x_p\!\mid\!x_g) &= s &
  \mathcal{L}_{-}(x_p\!\mid\!x_g) &= 1 - s\,.
\end{align}
When we backpropagate to either $x_g$ or $x_p$, we keep $\varphi_p$ or $\varphi_g$ frozen, respectively.
Intuitively, $\mathcal{L}_{+}$ highlights pixels whose modification most \emph{reduces} similarity, \ie, regions supporting the perceived similarity between the pair, whereas $\mathcal{L}_{-}$ emphasizes pixels whose change most \emph{increases} similarity, revealing evidence for current dissimilarity.
We conjecture, though, that defining dissimilar regions does not make sense in all cases; for example, it is difficult to manually analyze which regions in a non-matching pair should be most dissimilar.
Thus, we show visual results for $e_-$ as a reference, but focus our analysis on $e_+$.

%% file: sections/experiments.tex
\section{Experiments}
\label{sec:exp}

To demonstrate the versatility of our proposed FAME method across different visual understanding tasks, we conduct experiments on both IC and FR datasets.
For IC, we employ the ImageNet validation set \cite{deng2009imagenet}, which contains 50,000 images from 1,000 object categories, providing diverse visual concepts and well-defined bounding boxes, allowing quantitative evaluation of attribution maps.
In our experiments, we select 5 images for each of the 1000 classes that are correctly classified by all employed networks, and back-propagate the logits $z_o$ of the ground truth class.

For the FR task, we use three representative datasets: AR Face \cite{martinez1998arface}, CFP \cite{sengupta2016cfp}, and SCface \cite{grgic2011scface}, since they allow for an unadulterated evaluation of specific conditions of occlusions, face pose and image resolution, in opposition to other benchmark datasets that mix these conditions \cite{lu2024corrrise}.
The AR Face dataset contains variations in occlusions caused by scarves covering the mouth and sunglasses covering both eyes, and we follow standard protocols \cite{pereira20228years}.
The CFP dataset consists of frontal and profile faces with yaw angles greater than 60°, where only one side of the face is visible, and we adopt the default protocols \cite{sengupta2016cfp}.
SCface is captured by indoor surveillance cameras under different distances and resolutions, where the default protocols \cite{wallace2012cross} allow systematic evaluation of distance effects.
We follow standard image preprocessing and facial alignment routines.

During experiments, we compare performance of different models.
For IC, we test three pretrained models: ResNet34, ResNet50, and ResNet101, as well as the old VGG19 network and the state-of-the-art ConvNeXt-Tiny variant from the PyTorch \cite{pytorch} model zoo.
For FR, to analyze the impact of network depth on interpretability we adopt IResNet101\footnote{\url{https://github.com/mk-minchul/AdaFace}} along with two smaller variants, IResNet18 and IResNet50 \cite{kim2022adaface}, all pre-trained with AdaFace.

Moreover, we compare our method with several representative XAI techniques, including Grad-CAM \cite{selvaraju2017gradcam} and Grad-CAM-Elementwise (Grad-CAM-EW).
For IC, we additionally compare against HiResCAM \cite{draelos2020hirescam} and FullGradCAM \cite{srinivas2019fullgrad}, for FR we add CorrRISE \cite{lu2024corrrise} and FGGB \cite{lu2024fggb}, to demonstrate the effectiveness and generalizability of FAME across both tasks.
For CorrRISE \cite{lu2024corrrise}, we follow most of the implementation details described in their paper.
Specifically, 500 random masks are generated for each image pair, where each mask contains 10 black patches with the size of $30\times30$ that are multiplied with the original gallery and probe images to replace the corresponding pixels separately.
Since the official implementation is not publicly available, we reproduce FGGB strictly following the procedural details described in the original paper \cite{lu2024fggb}.
%
For FAME, we rely on the original LOTS parameters \cite{rozsa2017lots}, \ie, we use $\eta=1/255$ and 500 iterations.
For computing the FAME attribution, we use a Gaussian blur with standard deviation of 7.7, see \secref{sec:blur} for a discussion on this choice.

\subsection{How Reliable is CAM?}
\label{sec:reliable}

\input{figures/feature_maps}

As stated above, CAM-based visualization methods assume that elements in the feature map are only influenced by underlying pixel areas.
In this section, we use FAME via \eqref{eq:fame-fm} to falsify this assumption.
\subfigref{fig:activations-fm-img} reveals that different spatial positions in the feature map correspond to regions of varying size and coverage in the input image.
While some locations focus on localized object parts (\eg, the bear's head or ear), others capture information from much broader areas, extending across the body and surrounding background.
This observation confirms that the receptive fields of feature-map elements are neither uniform nor localized.
Some units integrate context from large regions of the input image rather than from spatially fixed patches.
Thus, interpreting CNNs through upsampled activation-level maps, as in CAM-based methods, can lead to misleading attribution.

To further verify this phenomenon, we apply FAME on FR networks, where we systematically examine how feature-map activations correspond to different input regions across architectures of varying depth.
As shown in \subfigref{fig:activations-fm-cfp}, the shallow network (IResNet34) tends to extract information from relatively fixed local areas, \ie, each spatial unit in the feature map corresponds to a specific region in the input.
In contrast, receptive fields of deeper networks (IResNet50 and IResNet101) expand and shift, causing activations at the top positions of the feature map to aggregate information from central regions of the input face rather than spatially aligned areas.
FAME provides more reliable spatial interpretation, since the elements of feature maps are not fixed representations of input regions, especially in deeper networks where receptive fields become large and overlapping.
It demonstrates that deeper architectures encode more abstract, globally integrated patterns, flagging CAM-based methods as unreliable.

\input{figures/ImageNET_diff_networks.tex}

\input{figures/fr_3_compact}

\subsection{Image Classification}

\input{figures/tab_eval_img}

\noindent\textbf{Visual Results.}
%
\figref{fig:activations-om-img-diff} qualitatively compares saliency maps generated by Grad-CAM-EW and FAME across different network architectures, including ResNet variants, VGG19, and ConvNeXt-Tiny, on two ImageNet samples.
For the bird, Grad-CAM-EW often produces diffuse responses that extend to the surrounding context, such as grass, or concentrates on the bird's head only, whereas FAME highlights regions concentrated around visually salient locations across the bird, across different backbones.
In the elephant example, Grad-CAM-EW typically activates broad areas mostly covering the head, while FAME emphasizes more compact regions around the entire body.
Notably, attribution patterns of FAME are similar across all networks.
These qualitative results indicate that FAME can be applied consistently across diverse deep network architectures, producing comparable sensitivity patterns. 

\noindent\textbf{Evaluation Metrics.}
To assess the quality of the explanations produced by FAME, we employ two complementary evaluation metrics.
Particularly, we adopt the Intersection over Union (IoU) metric as a localization proxy to quantify the spatial overlap between attribution maps and ground-truth object regions.
While IoU does not constitute a complete measure of explanation faithfulness, it remains a widely used evaluation metric in recent work \cite{li2025diffcam}.
We report IoU at multiple thresholds ($\texttt{thr} = 0.3, 0.5, 0.7$), where lower thresholds yield broader saliency and higher thresholds produce sparser, more discriminative maps.
This evaluation allows us to analyze the robustness of different XAI methods under varying levels of strictness in saliency selection.
Complementing IoU, we adopt ROAD-Delete \cite{rong2022road} as a removal-based faithfulness proxy, where removal ratio $P$ denotes the percentage of pixels removed according to the attribution ranking.
It evaluates how rapidly model performance degrades when the most salient regions are removed.
Consistent performance degradation across different $P$ values indicates that the explanation reliably captures regions that are critical to the model's decision.
While ROAD-Insert is used for our FR part later, we do not adopt it for IC, since object classification requires larger connected regions, whereas pixel-level attribution methods produce more localized and disjoint regions in IC (\cf~\figref{fig:activations-om-img-diff}), while FR relies on more localized features.

\noindent\textbf{Quantitative Evaluation.}
\tabref{tab:eval-img-r50} reports the quantitative comparison of different attribution methods on ImageNet.
In terms of IoU, FAME achieves competitive localization performance compared to CAM-based methods while remaining slightly below FullGradCAM across different thresholds.
This indicates that FAME provides reasonable spatial alignment with object regions.
However, not all parts of the object are important for classification, which shows conceptual limitations of the IoU metric for evaluation attribution.
When using ROAD-Delete to evaluate explanation quality through removal-based performance degradation, FAME achieves the highest scores at $P=10\%$ and $30\%$, and remains competitive at higher removal ratios.
This suggests that the most important regions highlighted by FAME are particularly relevant to the model prediction.

\input{figures/tab_eval_fr_all.tex}

\subsection{Face Recognition}
\noindent\textbf{Visual Results.}
\figref{fig:activations-fr} compares attribution maps from different XAI methods on three representative datasets using the largest IResNet101 model; additional backbones are shown in the supplemental material.
Grad-CAM averages channel weights across spatial locations (\cf~\cite{selvaraju2017gradcam}), which assumes feature map pooling.
Since FR networks do not perform such pooling (\cf~\secref{sec:method}), the resulting attributions appear spatially inconsistent.
Grad-CAM-EW highlights non-specific and spatially disjoint facial regions.
CorrRISE produces separated regions for $e_+$, while $e_-$ often corresponds to background.
These patterns remain noisy, particularly for small or low-resolution inputs in \subfigref{fig:fr-scface}.
FGGB decomposes the evidence into two components based on the threshold $\theta$ (cf.~\secref{sec:fggb}), selected at EER \cite{lu2024fggb,huber2024xssab}.
Although it highlights $e_+$ reasonably well, the spatial patterns are less stable across datasets, especially under occlusion or resolution degradation.
The largely missing $e_-$ visualization in \figref{fig:activations-fr} reflects its sensitivity to the threshold $\theta$.

FAME produces more spatially coherent and consistent attribution patterns across datasets.
On ARFace, FAME's $e_+$ concentrates on non-occluded identity cues such as the eyes and forehead, ignoring the covered mouth region, whereas $e_-$ highlights this region — for both gallery and probe images.
For SCFace, FAME remains robust to resolution degradation, with central activation near the nose and mouth.
On CFP, FAME captures the correspondence between frontal and profile views, focusing on shared structural areas like the nose bridge and cheek contour.
Interestingly, this holds for both the similar and dissimilar maps, which better follow our intuition than CorrRISE,\footnote{It is difficult to assess which facial regions are dissimilar. However, highlighting background that does not influence embeddings is incorrect.}  which arbitrarily highlights forehead or background regions.

\noindent\textbf{Evaluation Metrics.}
We follow the \emph{Deletion} and \emph{Insertion} strategy \cite{lu2024corrrise}, which measures how verification accuracy changes as salient pixels are progressively removed or inserted.
For each image pair, we compute the similarity score $s$ via \eqref{eq:cos}.
Given an attribution map $e_+$, we sort pixels by importance and iteratively delete or insert the top-$P$ percent ($P \in \{0,10,20,\ldots,100\}$) on the probe image $x_p$, while keeping the gallery image $x_g$ fixed to isolate the attribution effect.
After each perturbation, we recompute similarity and evaluate verification accuracy using the original decision threshold, yielding an accuracy-over-$P$ curve.
We summarize performance by the normalized area under this curve (AUC).
A faithful explanation produces a steeper accuracy drop (lower AUC) under deletion and a faster recovery (higher AUC) under insertion.

\noindent\textbf{Quantitative Evaluation.}
\tabref{tab:eval-fr-all-fggb} presents the quantitative comparison of different XAI methods evaluated using the Delete and Insert metrics across multiple FR protocols.
Overall, FAME consistently achieves competitive or superior scores compared with CAM-based approaches (Grad-CAM, Grad-CAM-EW), the perturbation-based CorrRISE, and the state-of-the-art FGGB.
In most protocols, FAME yields higher Insert values and comparable or lower Delete values, indicating that the regions identified by FAME are both highly informative and strongly aligned with the model's similarity computation.

On ARFace, FAME particularly improves under occluded conditions such as a glass and a scarf, suggesting better robustness in identifying the true discriminative regions when facial features are partially covered.
Similar improvements are observed on SCface, where FAME demonstrates more stable performance across varying camera distances, reflecting stronger generalization.
On CFP, FAME achieves the highest Insert scores (up to $96.7\%$), confirming its effectiveness under pose variations.
Compared to CAM-based methods, which are often affected by feature-map misalignment, cf.~\secref{sec:reliable}, FAME provides more consistent activation relevance by directly modeling the relationship between input regions and the similarity.

\begin{figure}[!b]
  \centering
  \subfloat[Deletion]{\label{fig:cfp-fp-r100-del}
    \includegraphics[page=21,width=.49\linewidth,trim=0 0 0 10mm,clip]{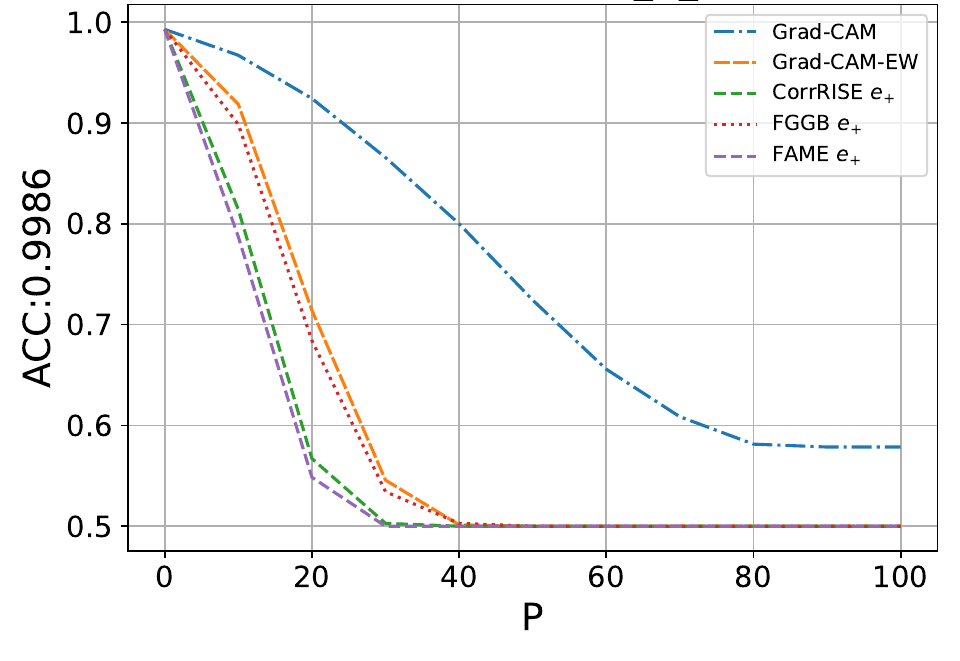}
 }%
  \subfloat[Insertion]{\label{fig:cfp-fp-r100-insert}
    \includegraphics[page=21,width=.49\linewidth,trim=0 0 0 10mm,clip]{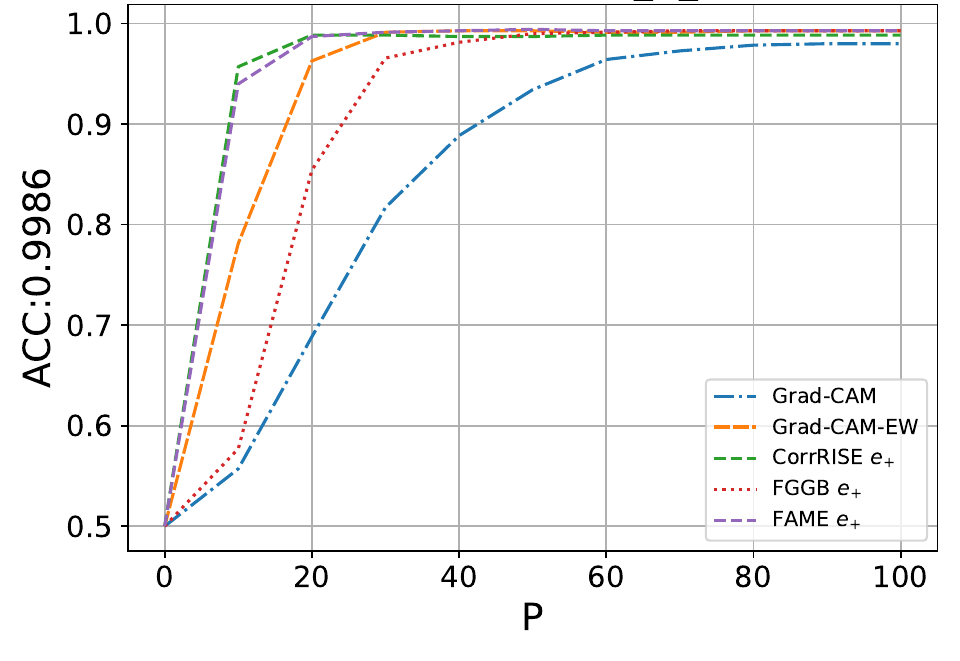}
 }
  \Caption[fig:cfp-fp-r100-eval]{Deletion vs. Insertion}{
 This figure shows \subref*{fig:cfp-fp-r100-del} Deletion and \subref*{fig:cfp-fp-r100-insert} Insertion evaluation curves for different XAI methods on the CFP \texttt{FP} protocol using IResNet101, which has a clean accuracy of 99.86\,\%.}
\end{figure}

\figref{fig:cfp-fp-r100-eval} compares the Deletion and Insertion curves of different XAI methods on the CFP \texttt{FP} protocol using the IResNet101 model.
In \subfigref{fig:cfp-fp-r100-del}, accuracy decreases as the most salient pixels are progressively removed from the probe image, while in the \subfigref{fig:cfp-fp-r100-insert}, accuracy increases as pixels are gradually added back.
As shown, FAME consistently outperforms CAM-based methods, exhibiting the fastest performance degradation under Deletion and the fastest accuracy gain by Insertion.
While CorrRISE has a slightly steeper increase for small $P$ in \subfigref{fig:cfp-fp-r100-insert}, indicating the most important regions are highlighted better, FAME surpasses CorrRISE at higher $P$, showing that it captures all relevant pixels more reliably.
This demonstrates that FAME's attribution maps identify regions that are highly relevant to face verification.
%
These quantitative results, together with the qualitative analysis in \figref{fig:activations-fr}, show that FAME produces visually meaningful explanations and better correlates with similarity scores across network depths and datasets.

\subsection{Effect of Gaussian Blur}
\label{sec:blur}

\input{figures/ARface_sim_diff_gau_only_glass}
To visualize the effect of different Gaussian blur sizes, we evaluate the IResNet101 model on one ARFace sample.
As shown in \figref{fig:fame-arface-gau-glass}, the qualitative structure of the attribution maps evolves smoothly with the degree of smoothing.
When the kernel size is small, as in \subfigref{fig:fm-a}, the maps retain fine-grained, high-frequency responses reflecting the raw perturbation field.
Although such maps appear noisy, the central activations are around the nose tip and mouth region. 
As the blur increases to moderate values in \subfigref{fig:fm-c}, the explanations become spatially coherent while preserving the main evidence regions.
This setting provides a good balance between visual smoothness and localization fidelity: the highlighted areas correspond to key identity cues (nose, mouth, and cheek contours) while suppressing noisy fluctuations and irrelevant activations on the forehead or chin.
In contrast, large smoothing in \subfigref{fig:fm-f} produces overly diffuse maps where almost the entire face is highlighted, leading to loss of spatial contrast and interpretive precision.

Overall, the results demonstrate that Gaussian smoothing acts as an effective regularizer for FAME's perturbation field, suppressing artifacts without altering the fundamental explanatory structure.
The persistence of nose- and mouth-centered activations across all parameter choices further confirms that FAME captures robust and semantically meaningful evidence for identity similarity even under occlusion conditions, such as with sunglasses.
For increased comparability to CAM-based attribution maps, and consistent with \cite{lu2024fggb,huber2024xssab}, we used a blur with a kernel size of 7.7 as shown in \subfigref{fig:fm-e}.
This seems appropriate to the two tasks that we perform, \ie, IC on ImageNet, and FR.
However, other tasks might require more fine-grained visualization, for which smaller kernel sizes would be more appropriate.

%% file: figures/feature_maps.tex
\begin{figure}[!t]
  \newcommand\img[1]{\includegraphics[width=0.064\textwidth]{#1}}

    \subfloat[Image Classification]{\label{fig:activations-fm-img}
     \centering
    \begin{tabular}{@{\,}c@{\,}c@{\,}c@{\,}c@{\,}c@{\,}c@{\,}c@{\,}}
      \img{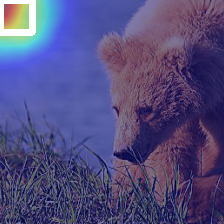} &
      \img{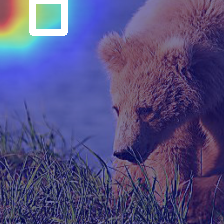} &
      \img{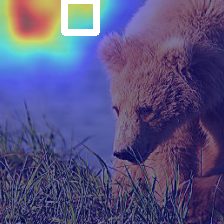} &
      \img{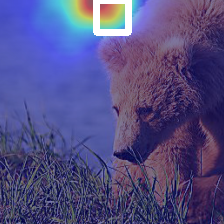}&
      \img{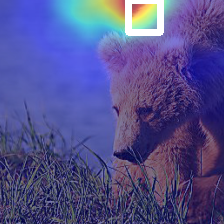}&
      \img{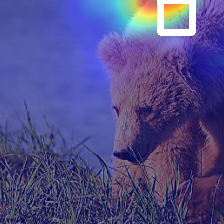}&
      \img{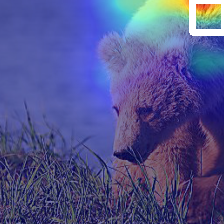}\\[-.5ex]

      \img{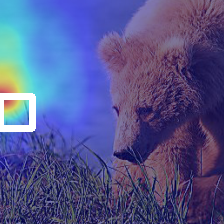} &
      \img{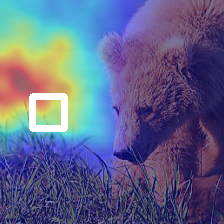} &
      \img{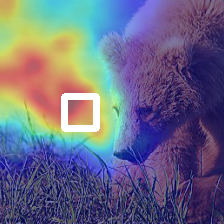} &
      \img{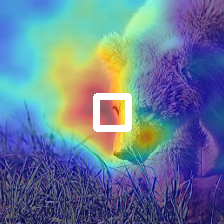}&
      \img{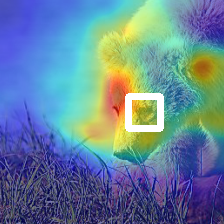}&
      \img{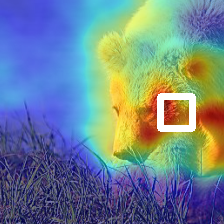}&
      \img{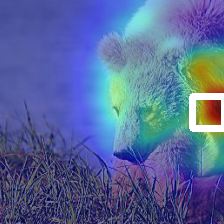}\\[-.5ex]

      \img{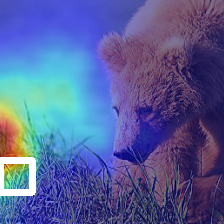} &
      \img{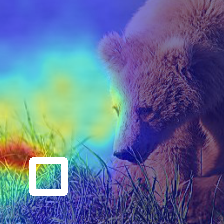} &
      \img{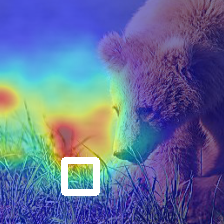} &
      \img{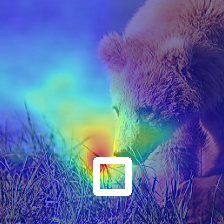}&
      \img{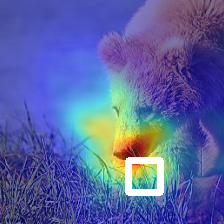}&
      \img{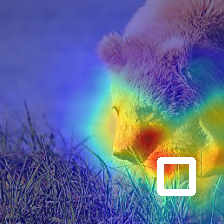}&
      \img{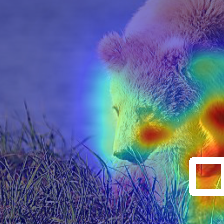}\\[-.5ex]

    \end{tabular}
  }

  \subfloat[Face Recognition]{\label{fig:activations-fm-cfp}
    \centering
    \begin{tabular}{@{\,}c@{\,}c@{\,}c@{\,}c@{\,}c@{\,}c@{\,}c@{\,}}
      \img{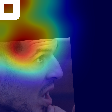} &
      \img{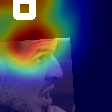} &
      \img{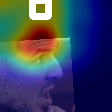} &
      \img{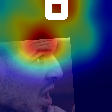}&
      \img{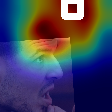}&
      \img{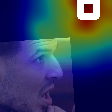}&
      \img{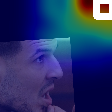}\\[-.5ex]

       \img{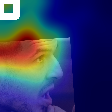} &
      \img{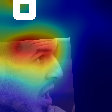} &
      \img{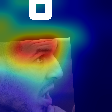} &
      \img{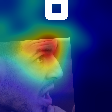}&
      \img{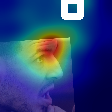}&
      \img{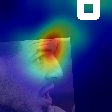}&
      \img{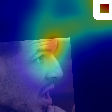}\\[-.5ex]

      \img{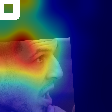} &
      \img{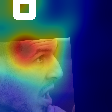} &
      \img{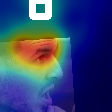} &
      \img{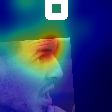}&
      \img{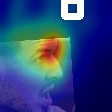}&
      \img{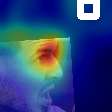}&
      \img{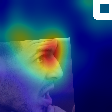}\\[-.5ex]

    \end{tabular}
  }

  \Caption[fig:activations-fm]{Feature Map Visualization}{
    The figure shows partial feature-map visualizations obtained using the proposed FAME method via $\mathcal L_a$ on the $7\times7$ feature map $a$.
    In \subref*{fig:activations-fm-img}, we show one exemplary ImageNet image, for rows 1, 4 and 6 of the feature map using ResNet101.
    In \subref*{fig:activations-fm-cfp}, we show the first row of features on a profile face image, using three networks IResNet34, IResNet50, and IResNet101.
    White borders indicate the pixels that lie underneath the particular feature map location $a[k]$.\\[-3ex]
  }

\end{figure}

%% file: figures/ImageNET_diff_networks.tex
\begin{figure*}
  \newcommand\img[1]{\includegraphics[width=0.064\textwidth]{#1}}
  \centering \small
    \begin{tabular}{@{}c@{\ \ }c@{\,}c@{\ \ }c@{\,}c@{\ \ }c@{\,}c@{\ \ }c@{\,}c@{\ \ }c@{\,}c@{}}
       & \multicolumn{2}{c}{ResNet34} & \multicolumn{2}{c}{ResNet50} & \multicolumn{2}{c}{ResNet101} & \multicolumn{2}{c}{VGG19} & \multicolumn{2}{@{}c@{}}{ConvNeXt\_Tiny} \\[-.5ex]
      \img{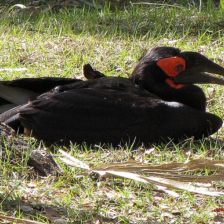} &

      \img{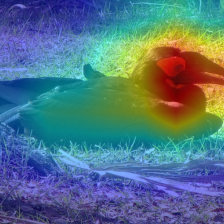} &
      \img{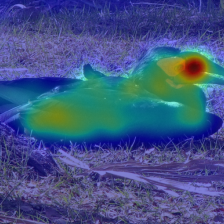} &

      \img{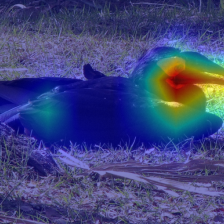} &
      \img{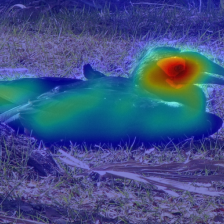} &

      \img{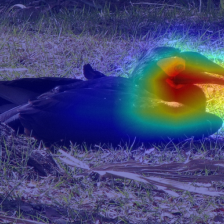} &
      \img{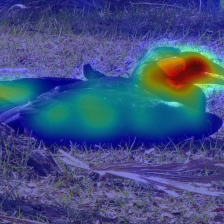}&

      \img{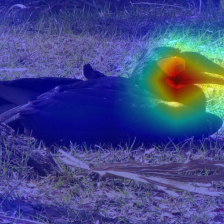} &
      \img{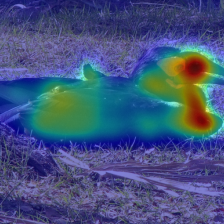} &

      \img{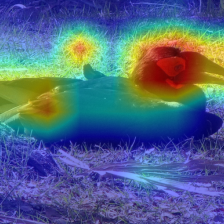} &
      \img{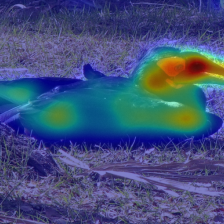}\\[-.3ex]


      \img{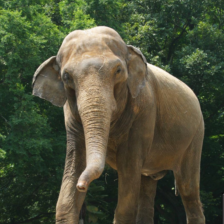} &

      \img{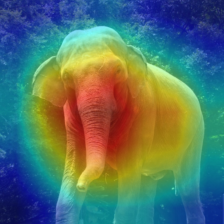} &
      \img{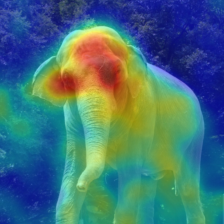} &

      \img{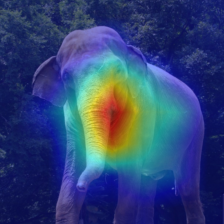} &
      \img{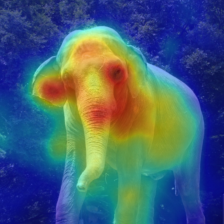} &
      \img{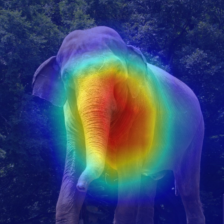} &
      \img{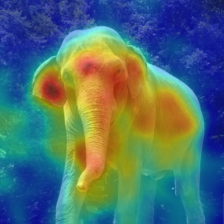}&

      \img{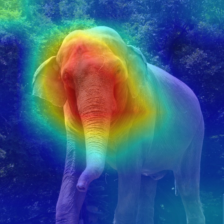} &
      \img{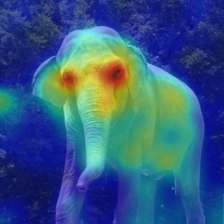}&

      \img{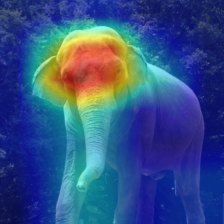} &
      \img{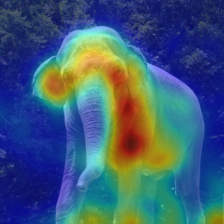}\\[-.5ex]
      \end{tabular}
  \Caption[fig:activations-om-img-diff]{ImageNet Visualization}{
    The figure shows the saliency maps generated by Grad-CAM-EW (left in each pair) and FAME (right) using different models on two ImageNet samples, evaluated with five different pre-trained networks.
  }

\end{figure*}

%% file: figures/fr_3_compact.tex
\begin{figure*}[!t]
  \newcommand\img[1]{\includegraphics[width=0.06\textwidth]{#1}}
  \centering\footnotesize
  \subfloat[\textbf{ARface} -- \texttt{scarf}]{\label{fig:fr-arface}
    \begin{tabular}{@{}c@{}c@{\ }c@{}c@{\ }c@{}c@{\,}c@{}c@{\ }c@{}c@{\,}c@{}c@{\ }c@{}c@{\,}c@{}c@{}}

      \multicolumn{2}{c}{Grad-CAM} &
      \multicolumn{2}{@{}c@{}}{Grad-CAM-EW} &
      \multicolumn{2}{c}{CorrRISE $e_+$} &
      \multicolumn{2}{c}{CorrRISE $e_-$} &
      \multicolumn{2}{c}{FGGB $e_+$} &
      \multicolumn{2}{c}{FGGB $e_-$} &
      \multicolumn{2}{c}{FAME $e_+$} &
      \multicolumn{2}{c}{FAME $e_-$}\\

      \img{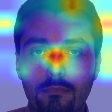} &
      \img{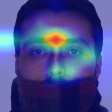} &
      \img{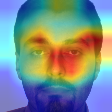} &
      \img{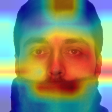}&
      \img{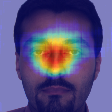}&
      \img{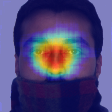}&
      \img{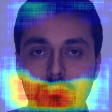}&
      \img{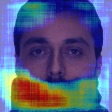}&
      \img{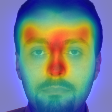}&
      \img{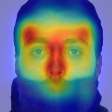}&
      \img{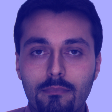}&
      \img{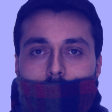}&
      \img{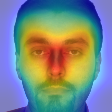}&
      \img{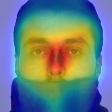}&
      \img{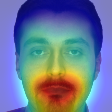}&
      \img{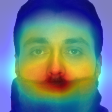}\\[-.5ex]
    \end{tabular}
  }

  \subfloat[\textbf{SCface} -- \texttt{medium}]{\label{fig:fr-scface}
    \begin{tabular}{@{}c@{}c@{\ }c@{}c@{\ }c@{}c@{\,}c@{}c@{\ }c@{}c@{\,}c@{}c@{\ }c@{}c@{\,}c@{}c@{}}
      \img{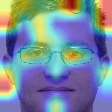} &
      \img{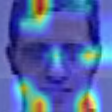} &
      \img{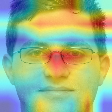} &
      \img{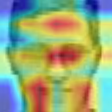}&
      \img{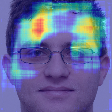}&
      \img{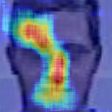}&
      \img{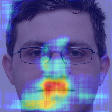}&
      \img{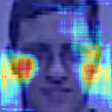}&
      \img{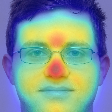}&
      \img{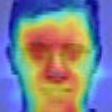}&
      \img{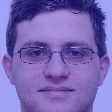}&
      \img{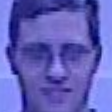}&
      \img{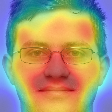}&
      \img{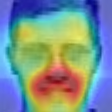}&
      \img{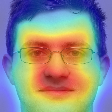}&
      \img{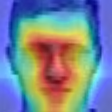}\\
    \end{tabular}
  }

  \subfloat[\textbf{CFP} -- \texttt{FP}]{\label{fig:fr-cfp}
    \begin{tabular}{@{}c@{}c@{\ }c@{}c@{\ }c@{}c@{\,}c@{}c@{\ }c@{}c@{\,}c@{}c@{\ }c@{}c@{\,}c@{}c@{}}
      \img{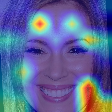} &
      \img{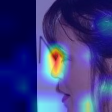} &
      \img{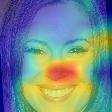} &
      \img{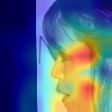}&
      \img{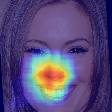}&
      \img{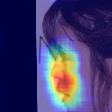}&
      \img{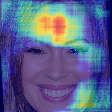}&
      \img{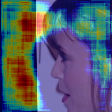}&
      \img{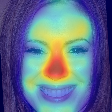}&
      \img{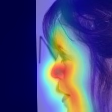}&
      \img{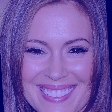}&
      \img{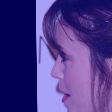}&
      \img{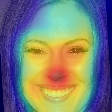}&
      \img{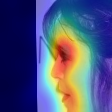}&
      \img{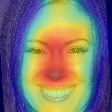}&
      \img{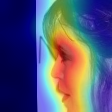}\\
    \end{tabular}
  }

  \Caption[fig:activations-fr]{Explaining Face Recognition}{
    The figure provides a comparative visualization of explanation maps generated on three representative datasets for genuine (same-identity) image pairs using IResNet101. XAI techniques include Grad-CAM, Grad-CAM-EW, CorrRISE, FGGB, and FAME, including similar $e_+$ and dissimilar attribution $e_-$ where appropriate.
  }

\end{figure*}

%% file: figures/tab_eval_img.tex

\begin{table}
  \Caption[tab:eval-img-r50]{Quantitative Evaluation of Attributions for Image Classification}{
    This table shows the quantitative comparison of different XAI methods on ImageNet.
    Higher IoU indicates better spatial alignment with ground-truth object regions, while higher ROAD-Delete scores indicate faster performance degradation when salient regions are removed.
    }
  \centering
  \footnotesize
  \setlength\tabcolsep{.75ex}
  \begin{tabular}{|l||c|c|c||c|c|c|}
    \hline
     & \multicolumn{3}{@{}c@{}||}{IoU in \% $\uparrow $} & \multicolumn{3}{c|}{ROAD-Delete $\uparrow$} \\[.3ex]
    \cline{2-7}
    XAI & thr=0.3 & thr=0.5 & thr=0.7 & P=10\% & P=30\% & P=50\% \\
    \hline
    Grad-CAM & 38.76 & 23.38 & 10.24 & \textit{0.2665} & \textit{1.0354} & 2.2562 \\ \hline
    Grad-CAM-EW & 40.15 & 24.40 & 10.70 & 0.2518 & 0.9749 & 2.1243 \\ \hline
    HiResCAM & 38.76 & 23.38 & 10.24  & \textit{0.2665} & 1.0354 & \textit{2.2563} \\ \hline
    FullGradCAM & \rednum{48.81} & \rednum{33.05} & \rednum{13.14} & 0.2527 & 1.0200 & \rednum{2.2888} \\ \hline
    FAME - $\mathcal L_{\mathrm{cls}}$ & \textit{46.09} & \textit{29.09} & \textit{10.79} & \rednum{0.4253} & \rednum{1.1499} & 2.1672  \\
    \hline
    \end{tabular}
\end{table}

%% file: figures/tab_eval_fr_all.tex
\begin{table*}
  \Caption[tab:eval-fr-all-fggb]{Quantitative Evaluation of Face Recognition Attribution}{
    This table compares different XAI methods evaluated using the Delete ($\downarrow$) and Insert ($\uparrow$) metrics across multiple FR protocols.
    Three AdaFace-based backbones (IResNet18, IResNet50, IResNet101) are tested on three datasets: AR Face, SCface, and CFP with default evaluation protocols.
    \textbf{Best} and \textit{second-best} results per model, protocol, and metric are highlighted.
  }
  \centering
  \footnotesize
  \setlength\tabcolsep{.6ex}
  \begin{tabular}{|l|c||c|c|c|c|c|c||c|c|c|c|c|c||c|c|c|c|}
    \hline
    && \multicolumn{6}{c||}{ARface} & \multicolumn{6}{c||}{SCface} & \multicolumn{4}{c|}{CFP}\\[.3ex]
    \cline{3-18}
    FR model& XAI&  \multicolumn{2}{c|}{neutral} & \multicolumn{2}{c|}{glass} & \multicolumn{2}{c||}{scarf} & \multicolumn{2}{c|}{close} & \multicolumn{2}{c|}{medium} & \multicolumn{2}{c||}{far}  & \multicolumn{2}{c|}{FF} &  \multicolumn{2}{c|}{FP}\\
    \cline{3-18}
    & & Delete & Insert & Delete & Insert & Delete & Insert & Delete & Insert & Delete & Insert & Delete & Insert & Delete & Insert & Delete & Insert \\\hline
     \multirow{5}{*}{IResNet18}
    & Grad-CAM & \textit{50.51} & 50.32 & 63.77 & 65.63 & 61.71 & 81.78 & 66.56 & 70.33 & 62.47 & 63.03 & 53.64 & 55.73 & 76.94 & 89.02 & 55.04 & 66.79 \\
    & Grad-CAM-EW & \rednum{50.47} & 52.48 & 54.35 & \textit{84.80} & 57.42 & 90.96 & 58.17 & \textit{85.90} & 55.95 & \rednum{78.74} & 51.88 & \rednum{60.90} & 64.51 & \textit{95.45} & 52.31 & \textit{78.54} \\
    & CorrRISE $e_+$ & \rednum{50.47} & 50.00 & 53.69 & 80.20 & \textit{53.84} & \textit{92.66} & \textit{55.87} & 84.23 & \textit{54.59} & 74.17 & \rednum{51.30} & 59.56 & 69.19 & 94.43 & \textit{51.94} & 77.90 \\
    & FGGB $e_+$ & \rednum{50.47} & \textit{52.69} & \rednum{52.85} & 75.93 & 54.10 & 85.44 & \rednum{55.83} & 80.13 & \rednum{53.74} & 72.91 & \textit{51.35} & 58.91 & \textit{62.89} & 88.10 & 52.19 & 73.31 \\
    & FAME $e_+$ & \rednum{50.47} & \rednum{52.78} & \textit{53.06} & \rednum{85.43} & \rednum{53.41} & \rednum{92.86} & 56.80 & \rednum{86.22} & 55.05 & \textit{78.53} & 52.40 & \textit{60.67} & \rednum{60.75} & \rednum{95.58} & \rednum{51.87} & \rednum{79.96} \\

    \hline \hline

    \multirow{5}{*}{IResNet50}
     & Grad-CAM & \textit{50.90} & 50.81 & 75.23 & 75.21 & 68.85 & 78.29 & 69.62 & 75.81 & 63.00 & 67.12 & 54.63 & 57.34 & 79.11 & 88.60 & 70.09 & 83.29 \\
     & Grad-CAM-EW & \rednum{50.77} & \textit{54.23} & 59.71 & 90.88 & 57.44 & 91.20 & 60.48 & 87.92 & 56.87 & \textit{78.17} & 52.64 & \textit{62.71} & 65.65 & 95.48 & 56.78 & 93.61 \\
     & CorrRISE $e_+$ & \rednum{50.77} & 50.00 & 59.55 & \textit{91.18} & \textit{55.07} & \textit{92.64} & 60.07 & \textit{88.62} & \textit{56.30} & 76.53 & \rednum{52.06} & 60.20 & 72.68 & \rednum{96.27} & \textit{55.69} & \textit{94.68} \\
     & FGGB $e_+$ & \rednum{50.77} & 54.02 & \textit{57.34} & 82.54 & 57.40 & 85.11 & \textit{59.59} & 82.78 & 56.48 & 72.01 & \textit{52.31} & 59.97 & \textit{64.78} & 88.36 & 57.55 & 88.91 \\
     & FAME $e_+$ & \rednum{50.77} & \rednum{55.00} & \rednum{56.85} & \rednum{91.75} & \rednum{54.46} & \rednum{92.84} & \rednum{58.31} & \rednum{88.94} & \rednum{56.17} & \rednum{78.97} & 53.43 & \rednum{63.14} & \rednum{62.20} & \textit{96.00} & \rednum{54.84} & \rednum{95.05} \\
    \hline \hline

    \multirow{5}{*}{IResNet101}
     & Grad-CAM & \textit{50.64} & 50.60 & 72.62 & 77.89 & 69.48 & 80.50 & 71.27 & 75.40 & 67.64 & 69.86 & 54.51 & 54.40 & 86.41 & 91.94 & 74.91 & 85.21 \\
     & Grad-CAM-EW & \rednum{50.60} & 52.91 & 60.44 & 90.09 & 59.12 & \textit{90.80} & 60.57 & \textit{87.17} & 59.05 & \textit{80.56} & 52.41 & \textit{59.48} & 70.17 & 96.70 & 59.26 & 94.36 \\
     & CorrRISE $e_+$ & \rednum{50.60} & 50.00 & 58.39 & \textit{91.24} & \textit{55.52} & \rednum{93.43} & \textit{58.59} & 86.78 & \textit{57.35} & 80.35 & \rednum{51.38} & 58.79 & 86.46 & \rednum{96.81} & \textit{56.31} & \textit{96.07} \\
     & FGGB $e_+$ & \rednum{50.60} & \textit{53.63} & \textit{58.09} & 83.71 & 58.17 & 86.61 & 59.59 & 83.17 & 58.49 & 75.58 & \textit{52.17} & 57.91 & \textit{69.10} & 91.39 & 58.66 & 90.85 \\
     & FAME $e_+$ & \rednum{50.60} & \rednum{54.06} & \rednum{56.12} & \rednum{92.07} & \rednum{55.13} & \rednum{93.43} & \rednum{58.10} & \rednum{88.22} & \rednum{57.19} & \rednum{82.02} & 52.73 & \rednum{60.15} & \rednum{65.70} & \textit{96.71} & \rednum{55.82} & \rednum{96.24} \\
    \hline

    \end{tabular}
\end{table*}

%% file: figures/ARface_sim_diff_gau_only_glass.tex
\begin{figure}[!t]
  \newcommand\img[1]{\includegraphics[width=1.3cm]{#1}}
  \centering
    \subfloat[$b=5$, $\sigma=0.8$]{\label{fig:fm-a}
    \begin{tabular}{@{}c@{}c@{\,}c@{}c@{\,}c@{}c@{}}

      \img{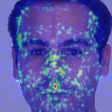} &
      \img{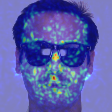}
       \end{tabular}
    }
    \subfloat[$b=11$, $\sigma=1.8$]{\label{fig:fm-b}
    \begin{tabular}{@{}c@{}c@{\,}c@{}c@{\,}c@{}c@{}}

      \img{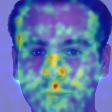} &
      \img{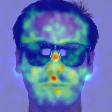}
       \end{tabular}
    }
    \subfloat[$b=15$, $\sigma=2.5$]{\label{fig:fm-c}
    \begin{tabular}{@{}c@{}c@{\,}c@{}c@{\,}c@{}c@{}}

      \img{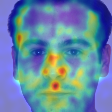} &
      \img{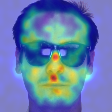}
       \end{tabular}
      }\\
    \subfloat[$b=25$, $\sigma=4$]{\label{fig:fm-d}
    \begin{tabular}{@{}c@{}c@{\,}c@{}c@{\,}c@{}c@{}}

      \img{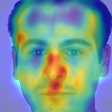} &
      \img{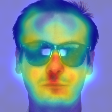}
       \end{tabular}
      }
    \subfloat[$b=49$, $\sigma=7.7$]{\label{fig:fm-e}
    \begin{tabular}{@{}c@{}c@{\,}c@{}c@{\,}c@{}c@{}}

      \img{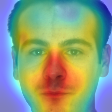} &
      \img{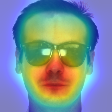}
       \end{tabular}
      }
    \subfloat[$b=81$, $\sigma=13$]{\label{fig:fm-f}
    \begin{tabular}{@{}c@{}c@{\,}c@{}c@{\,}c@{}c@{}}

      \img{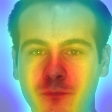} &
      \img{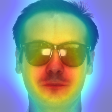}

    \end{tabular}
    }

  \Caption[fig:fame-arface-gau-glass]{Effect of Gaussian Smoothing}{
    This figure shows attribution maps produced by the proposed FAME method under different Gaussian blur parameters,
    where $b$ denotes the blur kernel size in pixels, and $\sigma$ the standard deviation of the Gaussian.
  }

\end{figure}

%% file: sections/conclusion.tex
\section{Conclusion}
\label{sec:conclusion}
In this work, we proposed our Feature Attribution Map Explanation (FAME) technique, a unified explainability framework that bridges gradient-based and perturbation-based paradigms for visual interpretation of deep neural networks.
Unlike conventional CAM-based methods that rely on a fixed spatial correspondence between feature maps and input regions, FAME explicitly models the pixel-level influence on intermediate activations through controlled gradient-driven perturbations.
It also removes the necessity to define thresholds as required by the most related FGGB method.
Comprehensive experiments on both IC and FR tasks demonstrate the versatility and effectiveness of the method, which works across various network topologies, including those performing pooling and flattening of the feature maps.
On ImageNet, FAME achieves competitive localization performance as measured by IoU, while emphasizing regions that are more relevant under deletion-based evaluation beyond the current state-of-the-art FullGradCAM.
On multiple FR datasets, FAME consistently outperforms Grad-CAM-EW, HiResCAM, CorrRISE and FGGB under both qualitative and quantitative evaluations, showing superior interpretation of occlusion, pose, and resolution variations, without requiring to select any threshold parameters to separate similar and dissimilar attribution maps.
Furthermore, our analysis on feature maps reveals that deeper networks capture information from increasingly large and shifted receptive fields, confirming that na\"ive spatial upsampling in CAM-based techniques is unreliable.

Overall, FAME provides a task-adaptive, optimization-based framework that yields stable and model-relevant sensitivity-based explanations across IC and FR in our experiments.
Future work will extend FAME to transformer-based architectures and other tasks, such as object detection, and explore its integration into human-centered explainability frameworks for trustworthy AI systems.
Notably, FAME is relatively slow due to the iterative gradient steps used to compute the adversarial sample via LOTS.
We used the default parameters proposed by \citet{rozsa2017lots}.
The supplemental material shows that larger step sizes $\eta$, fewer iterations, and loss-specific early stopping can increase speed with little impact on visualization quality.


%% file: sections/supplemental.tex
\section{Parameter Sensitivity and Optimization}
This section provides additional observations on how several hyperparameters influence the runtime and stability of our visualization procedure.
In particular, we examine the effect of larger step sizes and reduced iteration numbers.
These experiments show that faster configurations can often be used with minimal impact on the visual quality of the generated explanations.
\input{figures/ImageNET_diff_params}

The visualization results in \figref{fig:fame-parameters} show how different iteration numbers and step sizes influence the explanations produced by FAME when no early-stopping criterion is applied.
\subfigref{fig:fame-iterations} illustrates how the FAME visualization evolves across a wide range of iteration numbers.
With only 1 iteration, the explanation is extremely coarse and often lacks a meaningful localization on the object.
As the number of iterations increases, the heatmaps quickly become sharper and more focused, stabilizing between roughly 75 and 200 iterations.
Beyond 300 iterations, the changes are minor, and additional optimization mainly introduces more background activation rather than improving object localization.
This analysis highlights an important limitation of FGGB: the original FGGB method uses only a single step of backward propagation, which is insufficient to obtain a detailed or reliable attribution map, as our results clearly show.
In contrast, FAME benefits from multiple iterations, allowing it to refine the explanation and produce significantly more precise and interpretable visualizations.

Similarly, varying the step size in \subfigref{fig:fame-stepsize} affects the strength and smoothness of the activation: smaller $\eta$ yield smoother but weaker maps, while moderately larger step sizes produce stronger and more contrasted explanations.
Very large step sizes may introduce slight instability or noisy patterns in the background, although the main highlighted regions remain consistent.

\input{figures/tab_runtime}
\section{Runtime Evaluation}
\subtabref{tab:eval-img-runtime-r50} provides timing information obtained when visualizing 1000 images with a ResNet50 backbone.
The computation time increases approximately linearly with the number of iterations.
Even though larger iteration numbers lead to longer runtimes, our qualitative analysis indicates that meaningful visualizations can already be obtained with considerably fewer iterations, offering a practical trade-off between attribution quality and computational cost.
Loss-dependent early-stopping criteria might be applied to stop the iteration without losing visualization power.

In \subtabref{tab:eval-fr-runtime-cfp-fp-r100}, we further report the runtime comparison of all evaluated XAI techniques on face recognition attribution, using the 700 pairs contained in the CFP-FP protocol.
Grad-CAM and Grad-CAM-EW remain the fastest methods due to their single-pass backward computation that only goes back to the activation map.
CorrRISE and FGGB are significantly slower because they rely on repeated masking or iterative gradient computations for elements in the embeddings.
Our proposed FAME method with 500 iterations achieves a favorable balance between computational cost and attribution quality.
Although slower than Grad-CAM-based approaches, it is substantially faster than Corr\-RISE and FGGB while delivering the strongest quantitative performance.
This confirms that FAME provides an attractive trade-off, offering high-quality explanations with reasonable execution time for real-world verification analysis.

For a fair comparison with existing method implementations that are not parallelized, all visualizations as reported in \tabref{tab:runtime} are computed sequentially.
Our parallelized batch implementation can further increase speed when several visualizations need to be performed at the same time.

\section{Visual Results}
\input{figures/ImageNET-featuremaps}

\input{figures/CFP_features_maps_all}
\subsection{Feature Map Visualizations}
In Figure 1(a) of the main paper, we showed feature maps from one image classification network.
In \figref{fig:activations-fm-imagenet}, we show the complete results for all feature map elements for three different networks on ImageNet.
Similarly, in Figure 1(b) in the main paper, we showed only one row for visualizing the feature maps of a three face recognition networks.
\figref{fig:activations-fm-all} provides the remaining feature map locations.

\subsection{ImageNet Visualizations}

\input{figures/ImageNET_all.tex}
Figure 2 in the main paper showed some visual results for Grad-CAM-EW and FAME for different networks on ImageNet.
\figref{fig:activations-om-img-diff-supply} shows more examples.
As shown in \figref{fig:activations-om-img-diff-supply}, FAME produces spatially coherent and semantically focused attribution maps across different network architectures, including ResNet34, VGG19, and ConvNeXt-Tiny.
Compared with Grad-CAM, FullGradCAM, and HiResCAM, FAME consistently highlights the most discriminative object regions with clearer boundaries and reduced background noise.
Importantly, the qualitative patterns remain stable across architectures with substantially different design principles, ranging from classical convolutional networks (VGG19) to residual networks (ResNet34) and modern convolutional-transformer hybrids (ConvNeXt-Tiny).
This demonstrates that FAME does not rely on specific feature map structures and generalizes robustly across diverse backbone designs, yielding reliable and architecture-agnostic visual explanations.

\subsection{Face Recognition Visualization}
\input{figures/ARface_sim_3_networks}

\input{figures/SCface_sim_3_networks}
\input{figures/CFP_sim_3_networks}
Figure 3 in the main paper included face recognition examples from one network evaluated on three different datasets with one evaluation protocol each.
\figref{fig:activations-fr-arface}, \ref{fig:activations-fr-scface} and \ref{fig:activations-fr-cfp} include more networks and more protocols.

%% file: figures/ImageNET_diff_params.tex
\begin{figure*}[!t]
  \centering

  \newcommand\img[1]{\includegraphics[width=0.064\textwidth]{#1}}
  \subfloat[Iterations]{\label{fig:fame-iterations}
    \begin{tabular}{@{\,}c@{\,}c@{\,}c@{\,}c@{\,}c@{\,}c@{\,}c@{\,}c@{\,}c@{\,}c@{\,}}

     \centering
      \img{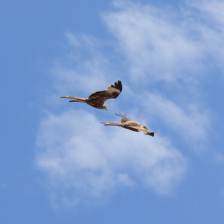} &
      \img{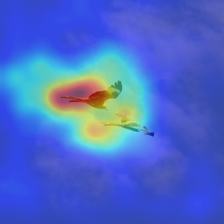} &
      \img{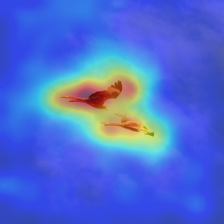} &
      \img{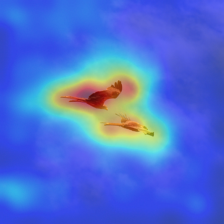} &
      \img{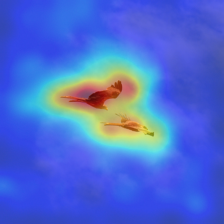} &
      \img{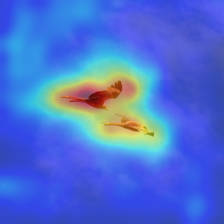} &
      \img{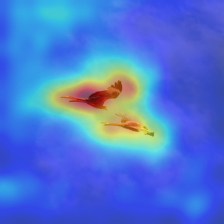} &
      \img{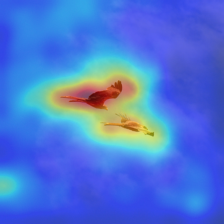} &
      \img{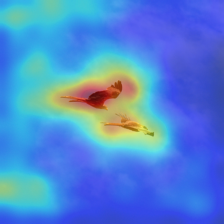} &
      \img{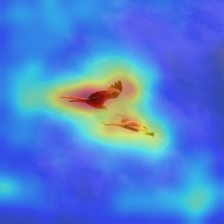}\\[-.5ex]

      \img{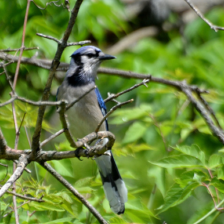} &
      \img{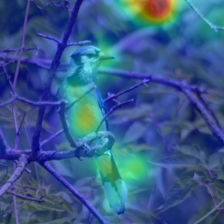} &
      \img{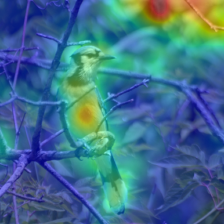} &
      \img{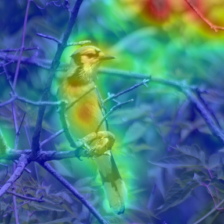} &
      \img{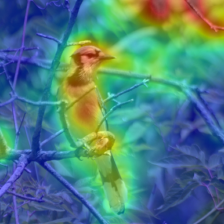} &
      \img{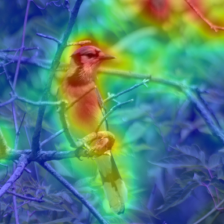} &
      \img{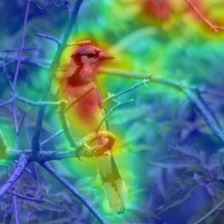} &
      \img{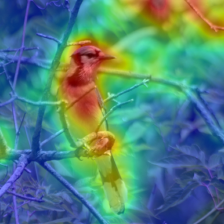} &
      \img{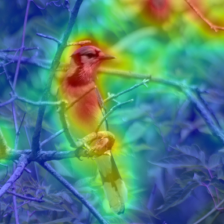} &
      \img{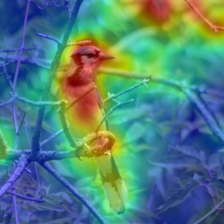}\\[-.5ex]

      \img{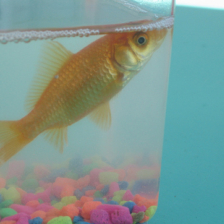} &
      \img{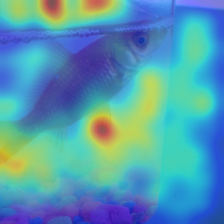} &
      \img{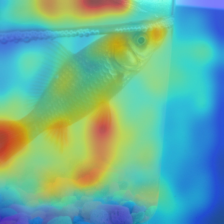} &
      \img{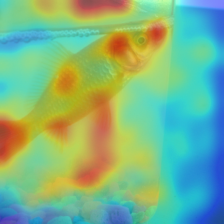} &
      \img{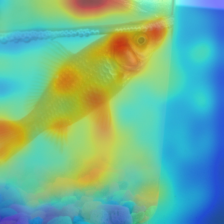} &
      \img{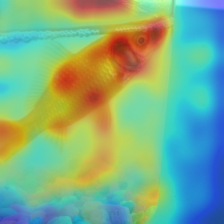} &
      \img{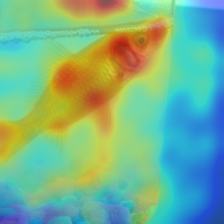} &
      \img{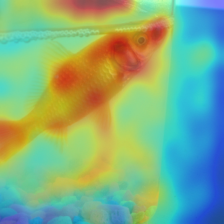} &
      \img{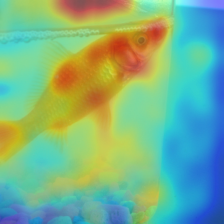} &
      \img{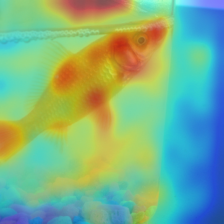}\\[-.5ex]
       &{$1$}&{$25$}&{$50$}&{$75$}&{$100$}&{$200$}&{$300$}&{$400$}&{$500$}\\

    \end{tabular}
  }

  \subfloat[Step Size $\eta$]{\label{fig:fame-stepsize}
    \begin{tabular}{@{\,}c@{\,}c@{\,}c@{\,}c@{\,}c@{\,}c@{\,}}

     \centering
      \img{images/ImageNet/single_img/ILSVRC2012_val_00027693} &
      \img{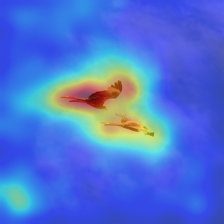} &
      \img{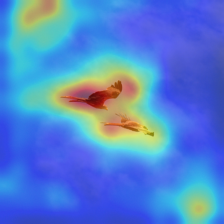} &
      \img{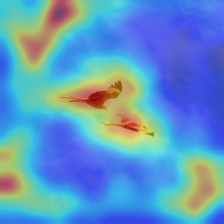} &
      \img{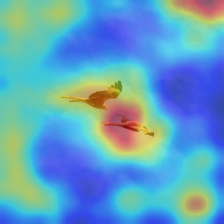} &
      \img{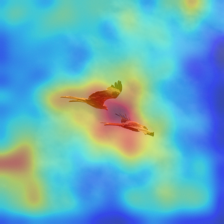}\\[-.5ex]

      \img{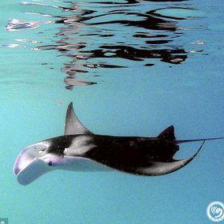} &
      \img{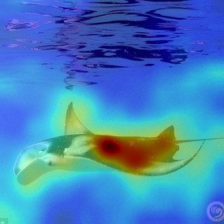} &
      \img{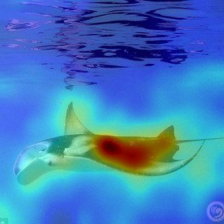} &
      \img{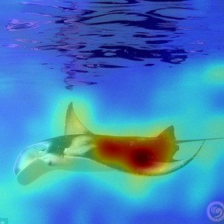} &
      \img{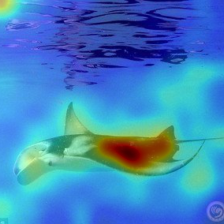} &
      \img{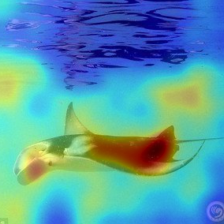}\\[-.5ex]

      \img{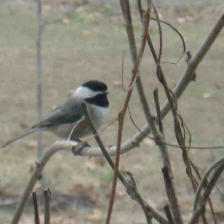} &
      \img{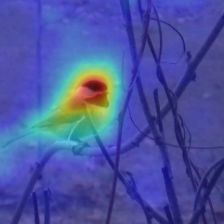} &
      \img{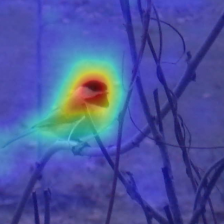} &
      \img{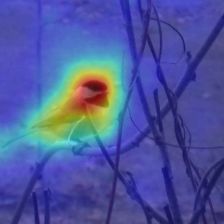} &
      \img{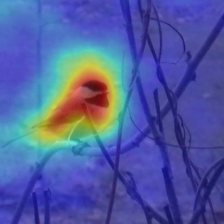} &
      \img{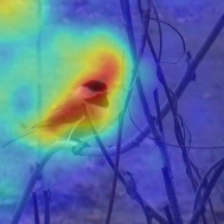}\\[-.5ex]
       &{$\frac{1}{255}$}&{$\frac{2}{255}$}&{$\frac{4}{255}$}&{$\frac{8}{255}$}&{$\frac{16}{255}$}\\

    \end{tabular}
    }

  \Caption[fig:fame-parameters]{Visualizations with FAME parameters}{
    The figure shows visualization results of the FAME method, when using \subref*{fig:fame-iterations} different numbers of iterations with a fixed step size $\eta=\frac{1}{255}$, and \subref*{fig:fame-stepsize} different step sizes $\eta$ with a fixed number of 100 iterations.
  }

\end{figure*}

%% file: figures/tab_runtime.tex
\begin{table}[t]
  \centering
  \Caption[tab:runtime]{Runtime Evaluation}{
    In \subref*{tab:eval-img-runtime-r50}, we provide the runtime of the FAME method for different numbers of optimization iterations using a ResNet50 model on 1000 ImageNet images.
    In \subref*{tab:eval-fr-runtime-cfp-fp-r100}, we compare the runtime of different XAI methods for face recognition attribution on the 700 image pairs contained in the CFP-FP protocol using the IResNet101 model. For fair comparison, all methods are run in a sequential manner.
    }
    \small
  \subfloat[FAME Iterations]{\label{tab:eval-img-runtime-r50}
    \begin{tabular}{|c|c|}
      \hline
      Iterations & Runtime (s) \\ \hline
      1 & 308.35\\ \hline
      25 & 733.95 \\ \hline
      50 & 1231.98 \\ \hline
      75 & 1678.72 \\ \hline
      100  & 2043.20 \\  \hline
      200  & 3975.97 \\  \hline
      300  & 5891.13 \\  \hline
      400  & 7626.18 \\  \hline
      500  & 9552.18  \\
      \hline
    \end{tabular}
  }
  \subfloat[Comparison]{\label{tab:eval-fr-runtime-cfp-fp-r100}
    \begin{tabular}{|c|c|}
      \hline
      XAI & Runtime (s) \\ \hline
      Grad-CAM & 366.15\\ \hline
      Grad-CAM-EW &455.55 \\ \hline
      CorrRISE & 7426.25 \\ \hline
      FGGB & 9558.98 \\ \hline
      FAME  & 2116.44  \\
      \hline
    \end{tabular}
  }
\end{table}

%% file: figures/ImageNET-featuremaps.tex
\begin{figure*}[p]
  \newcommand\img[1]{\includegraphics[width=0.064\textwidth]{#1}}
  \centering
  \subfloat[\textbf{ResNet34}]{\label{fig:imagenet-fm-resnet34}
    \begin{tabular}{@{\,}c@{\,}c@{\,}c@{\,}c@{\,}c@{\,}c@{\,}c@{\,}}
      \img{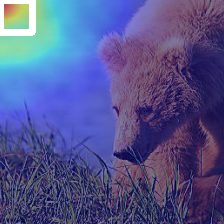} &
      \img{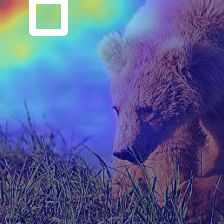} &
      \img{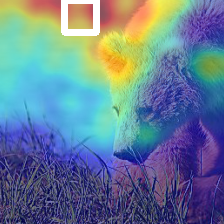} &
      \img{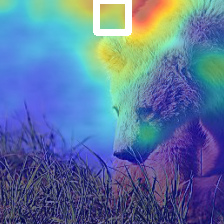} &
      \img{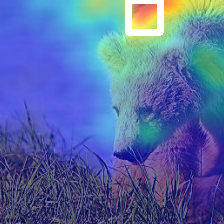} &
      \img{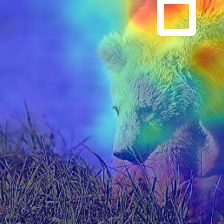} &
      \img{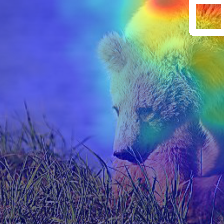} \\[-.5ex]

      \img{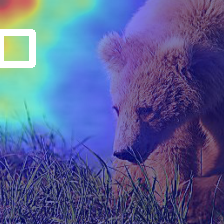} &
      \img{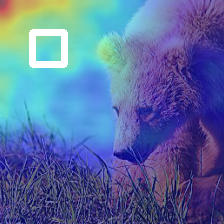} &
      \img{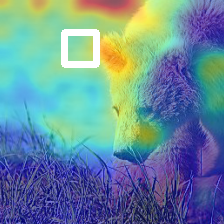} &
      \img{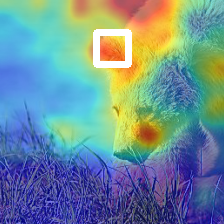} &
      \img{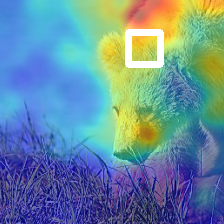} &
      \img{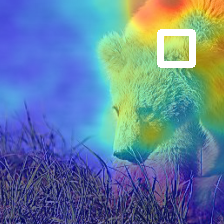} &
      \img{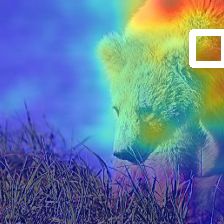} \\[-.5ex]

      \img{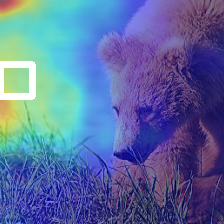} &
      \img{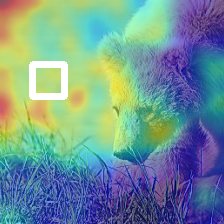} &
      \img{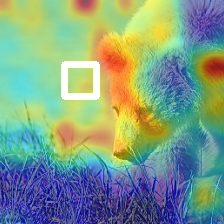} &
      \img{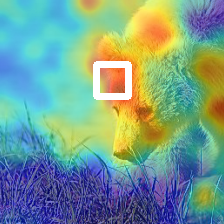} &
      \img{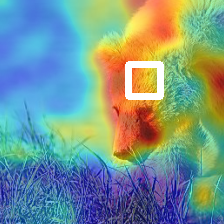} &
      \img{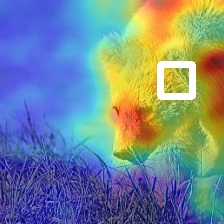} &
      \img{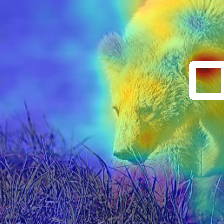} \\[-.5ex]

      \img{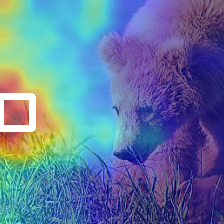} &
      \img{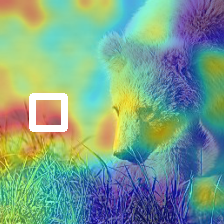} &
      \img{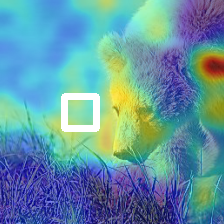} &
      \img{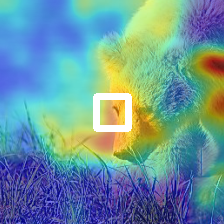} &
      \img{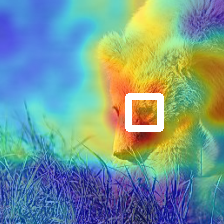} &
      \img{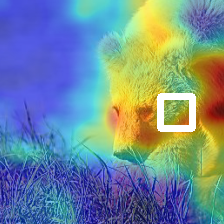} &
      \img{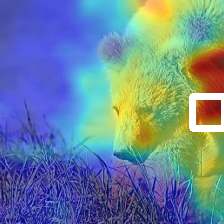} \\[-.5ex]

      \img{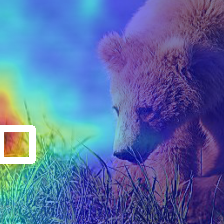} &
      \img{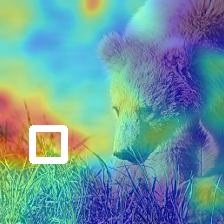} &
      \img{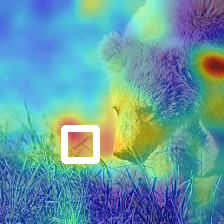} &
      \img{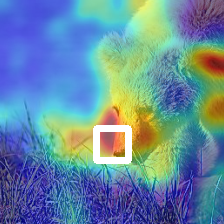} &
      \img{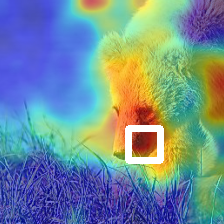} &
      \img{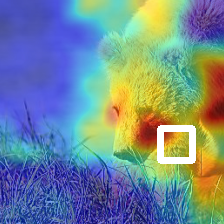} &
      \img{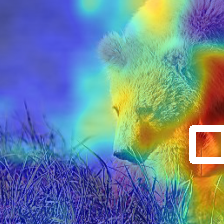} \\[-.5ex]

      \img{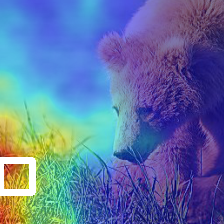} &
      \img{images/ImageNet/feature_map/ResNet34_ILSVRC2012_val_00010281.pdf_26} &
      \img{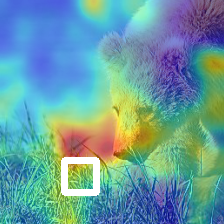} &
      \img{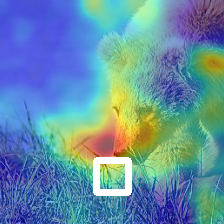} &
      \img{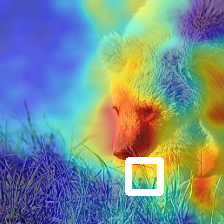} &
      \img{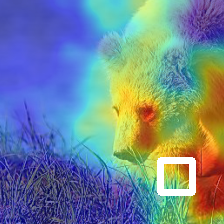} &
      \img{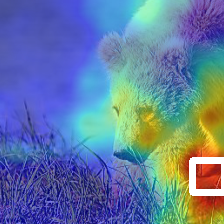} \\[-.5ex]

      \img{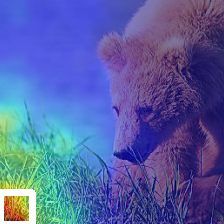} &
      \img{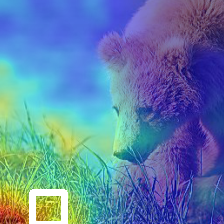} &
      \img{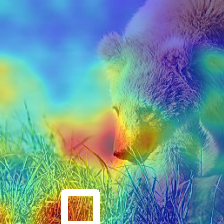} &
      \img{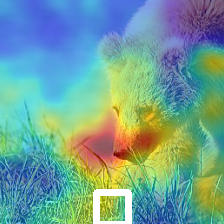} &
      \img{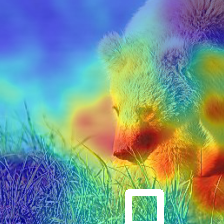} &
      \img{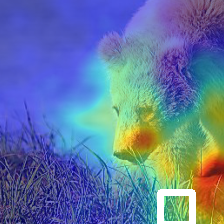} &
      \img{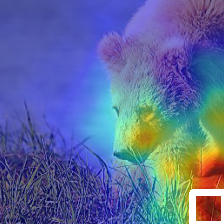}
    \end{tabular}
  }
  \subfloat[\textbf{ResNet50}]{\label{fig:imagenet-fm-resnet50}
    \begin{tabular}{@{\,}c@{\,}c@{\,}c@{\,}c@{\,}c@{\,}c@{\,}c@{\,}}
      \img{images/ImageNet/feature_map/ResNet50_ILSVRC2012_val_00010281.pdf_0} &
      \img{images/ImageNet/feature_map/ResNet50_ILSVRC2012_val_00010281.pdf_1} &
      \img{images/ImageNet/feature_map/ResNet50_ILSVRC2012_val_00010281.pdf_2} &
      \img{images/ImageNet/feature_map/ResNet50_ILSVRC2012_val_00010281.pdf_3} &
      \img{images/ImageNet/feature_map/ResNet50_ILSVRC2012_val_00010281.pdf_4} &
      \img{images/ImageNet/feature_map/ResNet50_ILSVRC2012_val_00010281.pdf_5} &
      \img{images/ImageNet/feature_map/ResNet50_ILSVRC2012_val_00010281.pdf_6} \\[-.5ex]

      \img{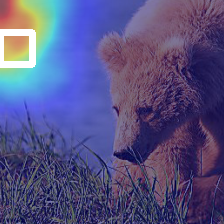} &
      \img{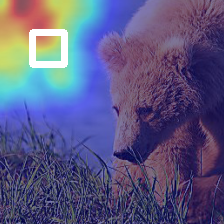} &
      \img{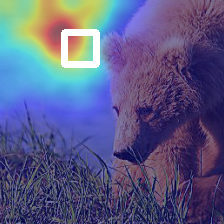} &
      \img{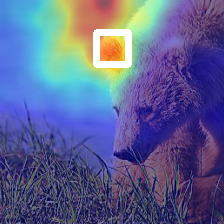} &
      \img{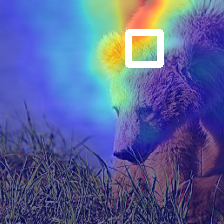} &
      \img{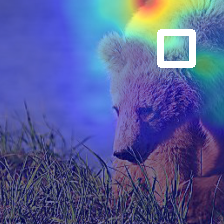} &
      \img{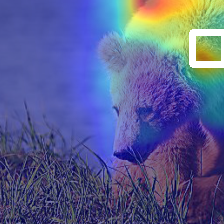} \\[-.5ex]

      \img{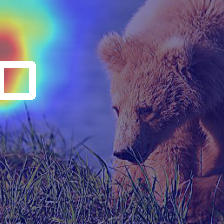} &
      \img{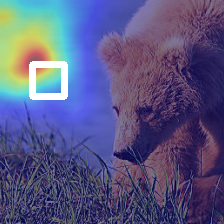} &
      \img{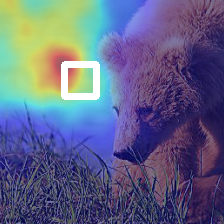} &
      \img{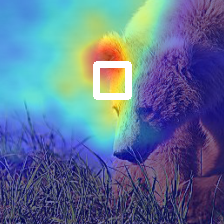} &
      \img{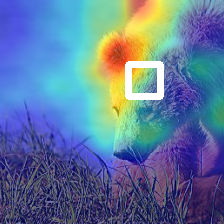} &
      \img{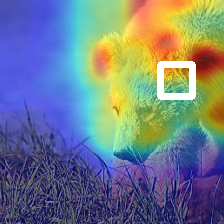} &
      \img{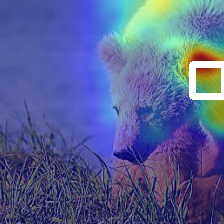} \\[-.5ex]

      \img{images/ImageNet/feature_map/ResNet50_ILSVRC2012_val_00010281.pdf_21} &
      \img{images/ImageNet/feature_map/ResNet50_ILSVRC2012_val_00010281.pdf_22} &
      \img{images/ImageNet/feature_map/ResNet50_ILSVRC2012_val_00010281.pdf_23} &
      \img{images/ImageNet/feature_map/ResNet50_ILSVRC2012_val_00010281.pdf_24} &
      \img{images/ImageNet/feature_map/ResNet50_ILSVRC2012_val_00010281.pdf_25} &
      \img{images/ImageNet/feature_map/ResNet50_ILSVRC2012_val_00010281.pdf_26} &
      \img{images/ImageNet/feature_map/ResNet50_ILSVRC2012_val_00010281.pdf_27} \\[-.5ex]

      \img{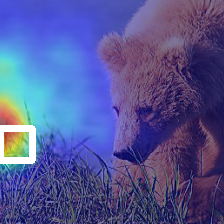} &
      \img{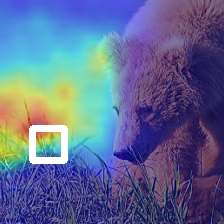} &
      \img{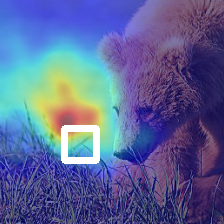} &
      \img{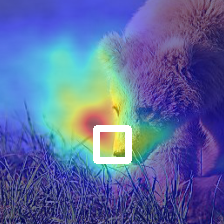} &
      \img{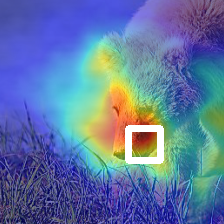} &
      \img{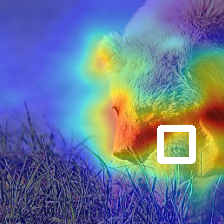} &
      \img{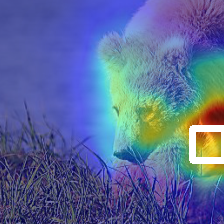} \\[-.5ex]

      \img{images/ImageNet/feature_map/ResNet50_ILSVRC2012_val_00010281.pdf_35} &
      \img{images/ImageNet/feature_map/ResNet50_ILSVRC2012_val_00010281.pdf_26} &
      \img{images/ImageNet/feature_map/ResNet50_ILSVRC2012_val_00010281.pdf_37} &
      \img{images/ImageNet/feature_map/ResNet50_ILSVRC2012_val_00010281.pdf_38} &
      \img{images/ImageNet/feature_map/ResNet50_ILSVRC2012_val_00010281.pdf_39} &
      \img{images/ImageNet/feature_map/ResNet50_ILSVRC2012_val_00010281.pdf_40} &
      \img{images/ImageNet/feature_map/ResNet50_ILSVRC2012_val_00010281.pdf_41} \\[-.5ex]

      \img{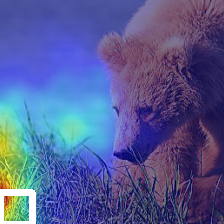} &
      \img{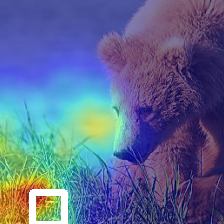} &
      \img{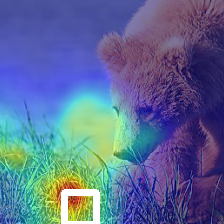} &
      \img{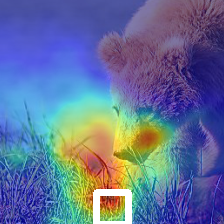} &
      \img{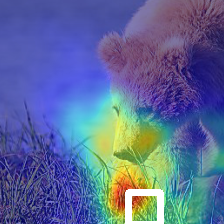} &
      \img{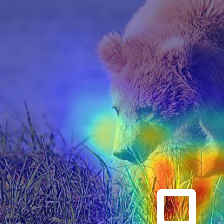} &
      \img{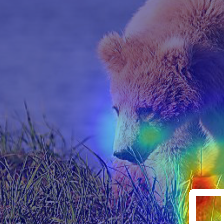}
    \end{tabular}
  }

  \subfloat[\textbf{ResNet101}]{\label{fig:imagenet-fm-resnet101}
    \begin{tabular}{@{\,}c@{\,}c@{\,}c@{\,}c@{\,}c@{\,}c@{\,}c@{\,}}
      \img{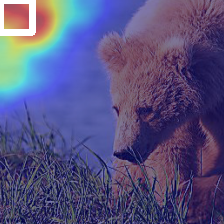} &
      \img{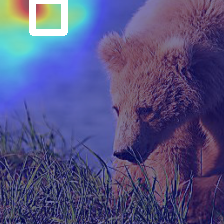} &
      \img{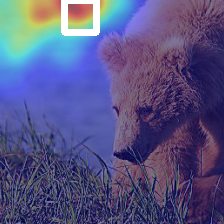} &
      \img{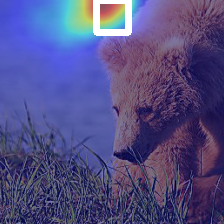} &
      \img{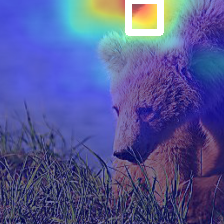} &
      \img{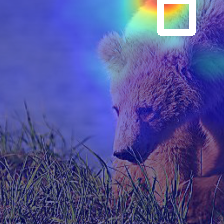} &
      \img{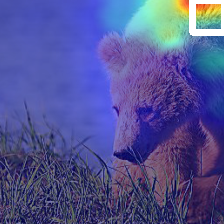} \\[-.5ex]

      \img{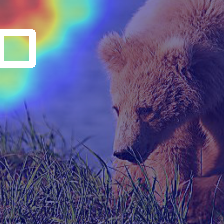} &
      \img{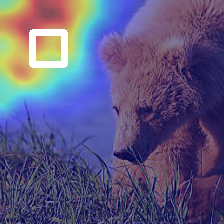} &
      \img{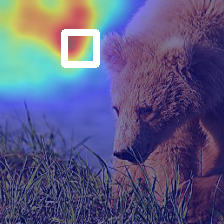} &
      \img{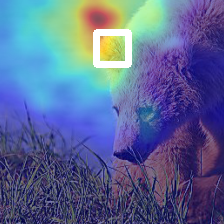} &
      \img{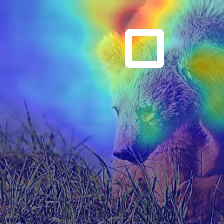} &
      \img{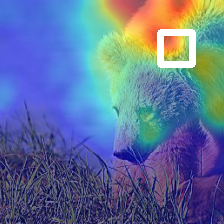} &
      \img{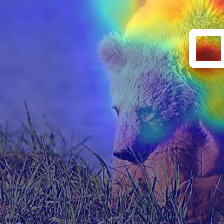} \\[-.5ex]

      \img{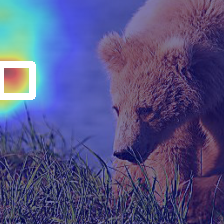} &
      \img{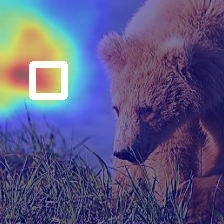} &
      \img{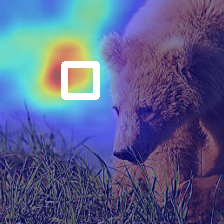} &
      \img{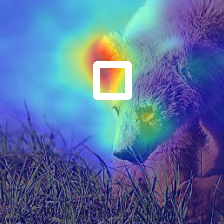} &
      \img{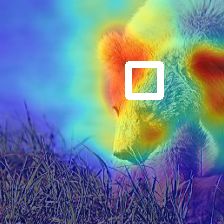} &
      \img{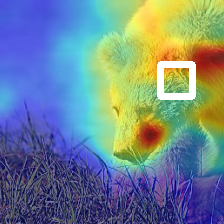} &
      \img{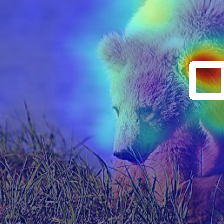} \\[-.5ex]

      \img{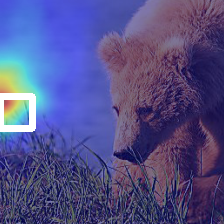} &
      \img{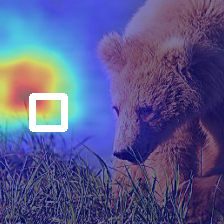} &
      \img{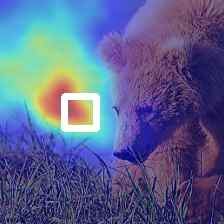} &
      \img{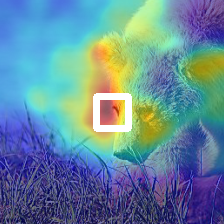} &
      \img{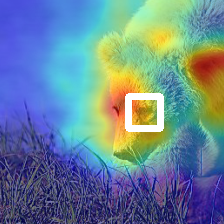} &
      \img{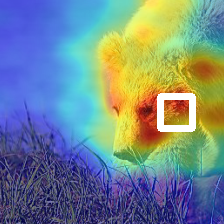} &
      \img{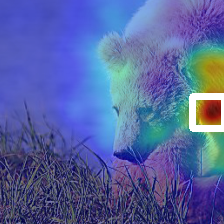} \\[-.5ex]

      \img{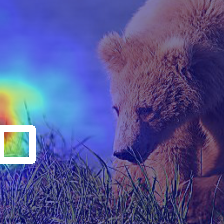} &
      \img{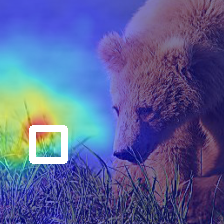} &
      \img{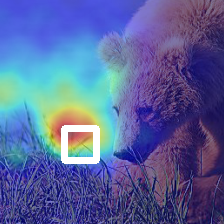} &
      \img{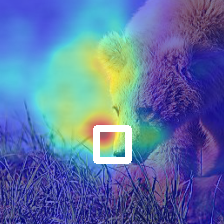} &
      \img{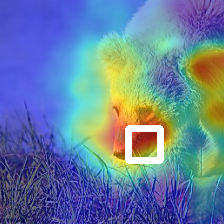} &
      \img{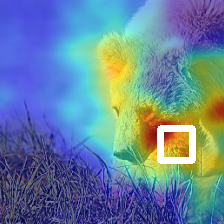} &
      \img{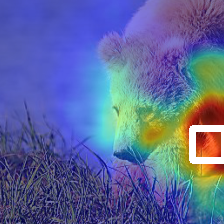} \\[-.5ex]

      \img{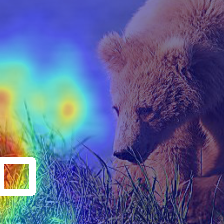} &
      \img{images/ImageNet/feature_map/ResNet101_ILSVRC2012_val_00010281.pdf_26} &
      \img{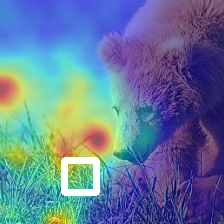} &
      \img{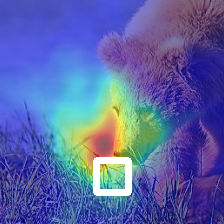} &
      \img{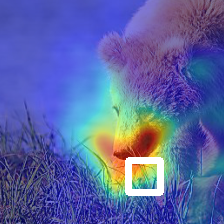} &
      \img{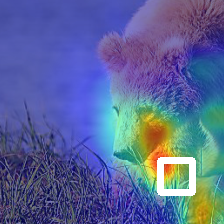} &
      \img{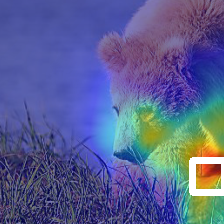} \\[-.5ex]

      \img{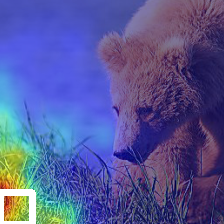} &
      \img{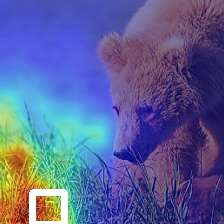} &
      \img{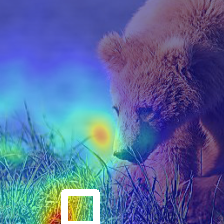} &
      \img{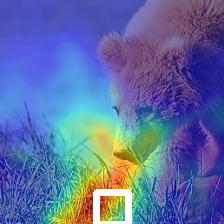} &
      \img{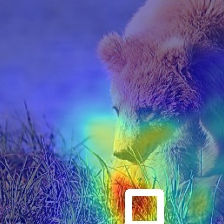} &
      \img{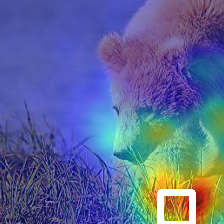} &
      \img{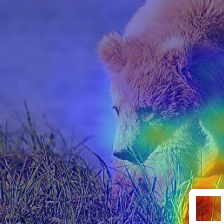}
    \end{tabular}
  }

  \Caption[fig:activations-fm-imagenet]{Feature Map Visualization for ImageNet Classification}{
    The figure shows FAME visualizations of the corresponding regions that are activated for all feature map locations in three image classification networks.
  }

\end{figure*}

%% file: figures/CFP_features_maps_all.tex
\begin{figure*}[p]
  \newcommand\img[1]{\includegraphics[width=0.064\textwidth]{#1}}
  \centering
  \subfloat[\textbf{IResNet18}]{\label{fig:cfp-fm-r18}
    \begin{tabular}{@{\,}c@{\,}c@{\,}c@{\,}c@{\,}c@{\,}c@{\,}c@{\,}}
      \img{images/CFP/feature_map/Adaface_ir_18/002_profile_02_feature_map_0} &
      \img{images/CFP/feature_map/Adaface_ir_18/002_profile_02_feature_map_1} &
      \img{images/CFP/feature_map/Adaface_ir_18/002_profile_02_feature_map_2} &
      \img{images/CFP/feature_map/Adaface_ir_18/002_profile_02_feature_map_3}&
      \img{images/CFP/feature_map/Adaface_ir_18/002_profile_02_feature_map_4}&
      \img{images/CFP/feature_map/Adaface_ir_18/002_profile_02_feature_map_5}&
      \img{images/CFP/feature_map/Adaface_ir_18/002_profile_02_feature_map_6}\\[-.5ex]

      \img{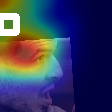} &
      \img{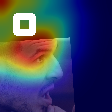} &
      \img{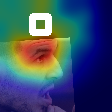} &
      \img{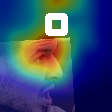}&
      \img{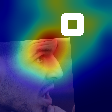}&
      \img{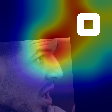}&
      \img{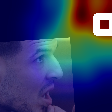}\\[-.5ex]

      \img{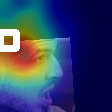} &
      \img{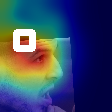} &
      \img{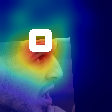} &
      \img{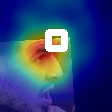}&
      \img{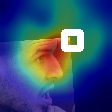}&
      \img{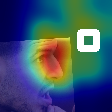}&
      \img{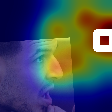}\\[-.5ex]

      \img{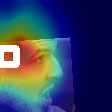} &
      \img{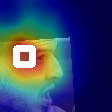} &
      \img{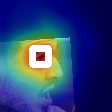} &
      \img{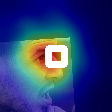}&
      \img{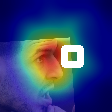}&
      \img{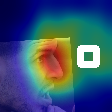}&
      \img{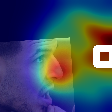}\\[-.5ex]

      \img{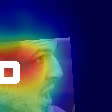} &
      \img{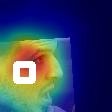} &
      \img{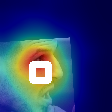} &
      \img{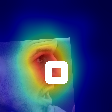}&
      \img{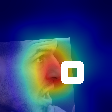}&
      \img{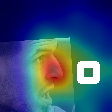}&
      \img{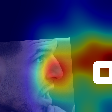}\\[-.5ex]
      \img{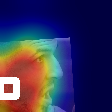} &
      \img{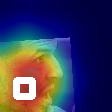} &
      \img{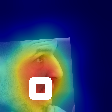} &
      \img{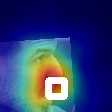}&
      \img{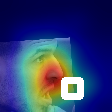}&
      \img{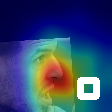}&
      \img{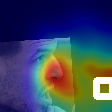}\\[-.5ex]

      \img{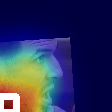} &
      \img{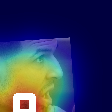} &
      \img{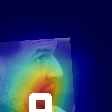} &
      \img{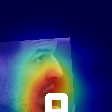}&
      \img{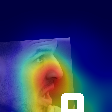}&
      \img{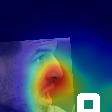}&
      \img{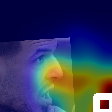}
    \end{tabular}
  }
  \subfloat[\textbf{IResNet50}]{\label{fig:cfp-fm-r50}
    \begin{tabular}{@{\,}c@{\,}c@{\,}c@{\,}c@{\,}c@{\,}c@{\,}c@{\,}}
      \img{images/CFP/feature_map/Adaface_ir_50/002_profile_02_feature_map_0} &
      \img{images/CFP/feature_map/Adaface_ir_50/002_profile_02_feature_map_1} &
      \img{images/CFP/feature_map/Adaface_ir_50/002_profile_02_feature_map_2} &
      \img{images/CFP/feature_map/Adaface_ir_50/002_profile_02_feature_map_3}&
      \img{images/CFP/feature_map/Adaface_ir_50/002_profile_02_feature_map_4}&
      \img{images/CFP/feature_map/Adaface_ir_50/002_profile_02_feature_map_5}&
      \img{images/CFP/feature_map/Adaface_ir_50/002_profile_02_feature_map_6}\\[-.5ex]

      \img{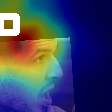} &
      \img{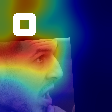} &
      \img{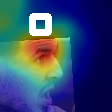} &
      \img{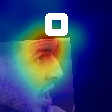}&
      \img{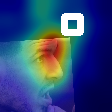}&
      \img{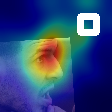}&
      \img{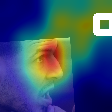}\\[-.5ex]

      \img{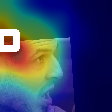} &
      \img{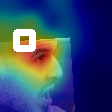} &
      \img{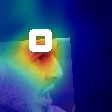} &
      \img{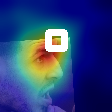}&
      \img{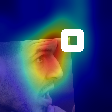}&
      \img{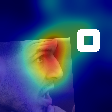}&
      \img{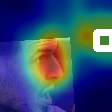}\\[-.5ex]

      \img{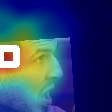} &
      \img{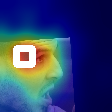} &
      \img{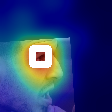} &
      \img{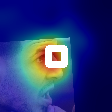}&
      \img{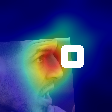}&
      \img{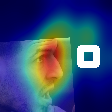}&
      \img{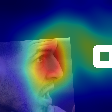}\\[-.5ex]

      \img{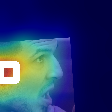} &
      \img{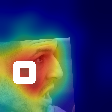} &
      \img{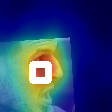} &
      \img{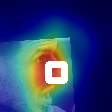}&
      \img{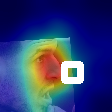}&
      \img{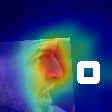}&
      \img{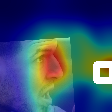}\\[-.5ex]
      \img{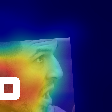} &
      \img{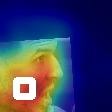} &
      \img{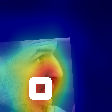} &
      \img{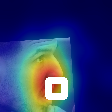}&
      \img{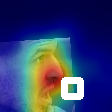}&
      \img{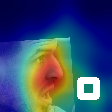}&
      \img{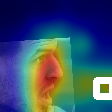}\\[-.5ex]

      \img{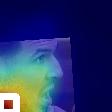} &
      \img{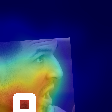} &
      \img{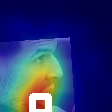} &
      \img{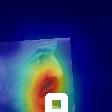}&
      \img{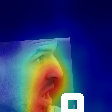}&
      \img{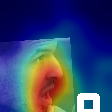}&
      \img{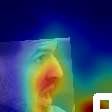}
    \end{tabular}
  }

  \subfloat[\textbf{IResNet101}]{\label{fig:cfp-fm-r101}
    \begin{tabular}{@{\,}c@{\,}c@{\,}c@{\,}c@{\,}c@{\,}c@{\,}c@{\,}}
      \img{images/CFP/feature_map/Adaface_ir_101/002_profile_02_feature_map_0} &
      \img{images/CFP/feature_map/Adaface_ir_101/002_profile_02_feature_map_1} &
      \img{images/CFP/feature_map/Adaface_ir_101/002_profile_02_feature_map_2} &
      \img{images/CFP/feature_map/Adaface_ir_101/002_profile_02_feature_map_3}&
      \img{images/CFP/feature_map/Adaface_ir_101/002_profile_02_feature_map_4}&
      \img{images/CFP/feature_map/Adaface_ir_101/002_profile_02_feature_map_5}&
      \img{images/CFP/feature_map/Adaface_ir_101/002_profile_02_feature_map_6}\\[-.5ex]

      \img{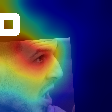} &
      \img{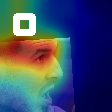} &
      \img{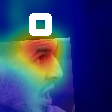} &
      \img{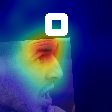}&
      \img{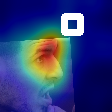}&
      \img{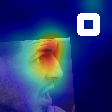}&
      \img{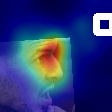}\\[-.5ex]

      \img{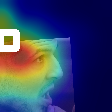} &
      \img{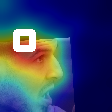} &
      \img{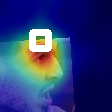} &
      \img{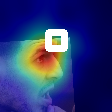}&
      \img{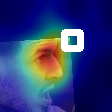}&
      \img{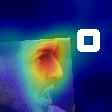}&
      \img{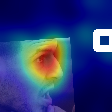}\\[-.5ex]

      \img{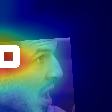} &
      \img{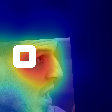} &
      \img{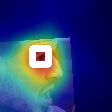} &
      \img{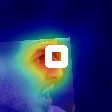}&
      \img{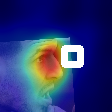}&
      \img{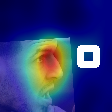}&
      \img{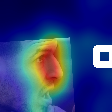}\\[-.5ex]

      \img{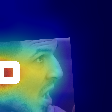} &
      \img{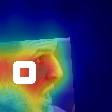} &
      \img{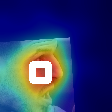} &
      \img{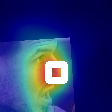}&
      \img{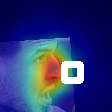}&
      \img{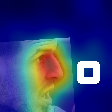}&
      \img{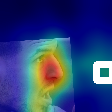}\\[-.5ex]
      \img{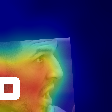} &
      \img{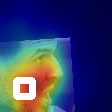} &
      \img{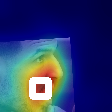} &
      \img{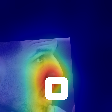}&
      \img{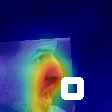}&
      \img{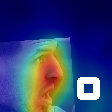}&
      \img{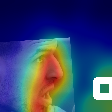}\\[-.5ex]

      \img{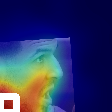} &
      \img{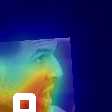} &
      \img{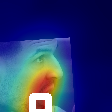} &
      \img{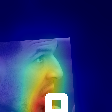}&
      \img{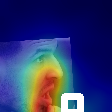}&
      \img{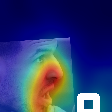}&
      \img{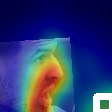}
      \end{tabular}
    }

  \Caption[fig:activations-fm-all]{Feature Map Visualization for Face Recognition}{
    The figure shows FAME visualizations of the corresponding regions that are activated for all feature map locations in three face recognition networks.
  }

\end{figure*}

%% file: figures/ImageNET_all.tex
\begin{figure*}[!b]
  \newcommand\img[1]{\includegraphics[width=0.07\textwidth]{#1}}
  \centering \small
    \begin{tabular}{@{}c@{\ }c@{}c@{}c@{}c@{\ }c@{}c@{}c@{}c@{\ }c@{}c@{}c@{}c@{}}
       & \multicolumn{4}{c}{ResNet34} & \multicolumn{4}{c}{VGG19} & \multicolumn{4}{c}{ConvNeXt\_Tiny} \\[-.5ex]
      \img{images/ImageNet/single_img/ILSVRC2012_val_00002100} &

      \img{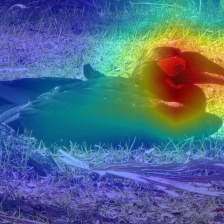} &
      \img{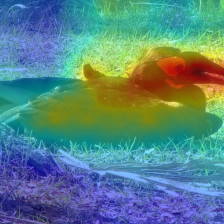} &
      \img{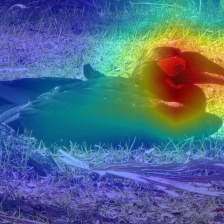} &
      \img{images/ImageNet/single_img/ResNet34_fame_ILSVRC2012_val_00002100} &

      \img{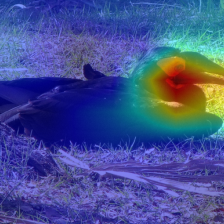} &
      \img{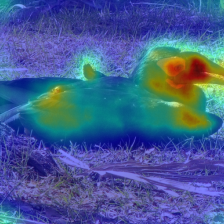} &
      \img{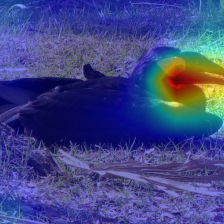} &
      \img{images/ImageNet/single_img/VGG19_fame_ILSVRC2012_val_00002100} &

      \img{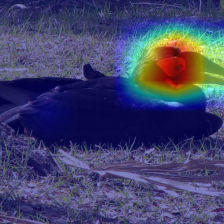} &
      \img{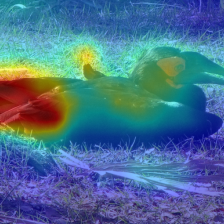} &
      \img{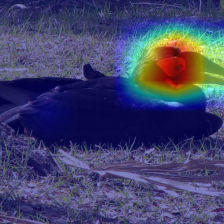} &
      \img{images/ImageNet/single_img/ConvNeXt_Tiny_fame_ILSVRC2012_val_00002100}\\[-.3ex]

      \img{images/ImageNet/single_img/ILSVRC2012_val_00013733} &

      \img{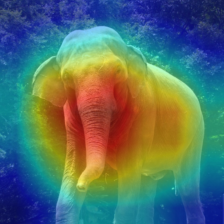} &
      \img{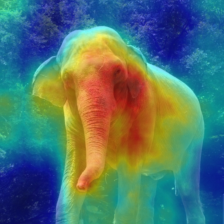} &
      \img{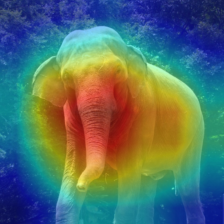} &
      \img{images/ImageNet/single_img/ResNet34_fame_ILSVRC2012_val_00013733} &

      \img{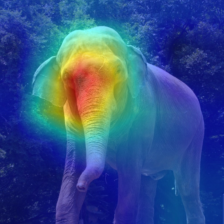} &
      \img{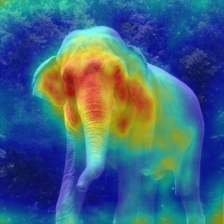} &
      \img{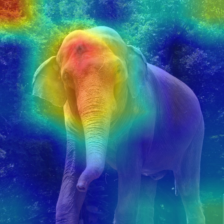} &
      \img{images/ImageNet/single_img/VGG19_fame_ILSVRC2012_val_00013733} &

      \img{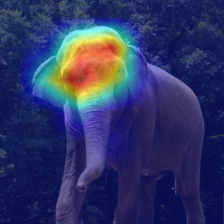} &
      \img{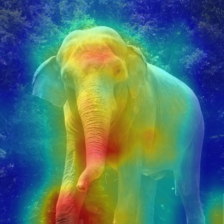} &
      \img{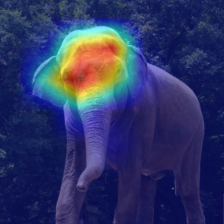} &
      \img{images/ImageNet/single_img/ConvNeXt_Tiny_fame_ILSVRC2012_val_00013733}\\[-.3ex]

       \img{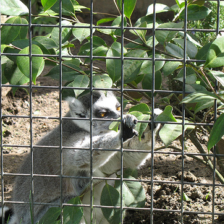} &

      \img{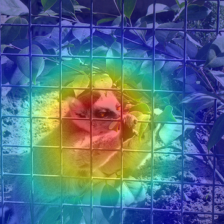} &
      \img{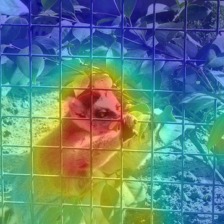} &
      \img{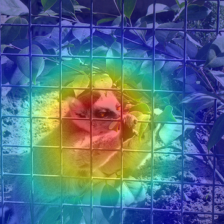} &
      \img{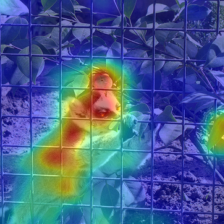} &

      \img{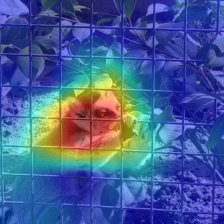} &
      \img{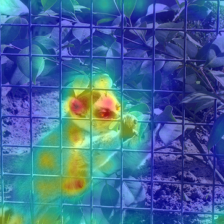} &
      \img{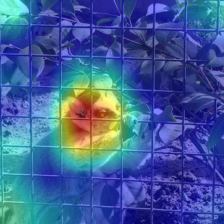} &
      \img{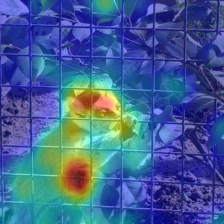} &

      \img{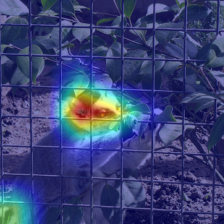} &
      \img{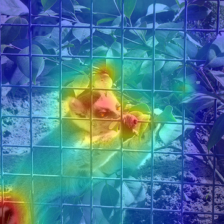} &
      \img{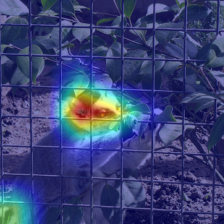} &
      \img{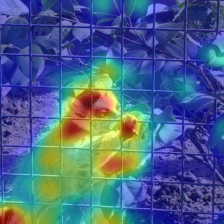}\\[-.3ex]

      \img{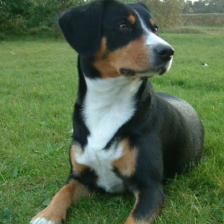} &
      \img{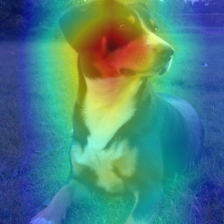} &
      \img{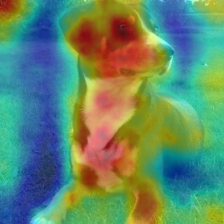} &
      \img{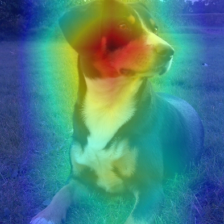} &
      \img{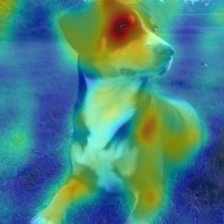} &

      \img{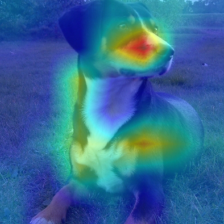} &
      \img{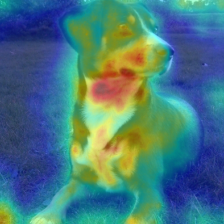} &
      \img{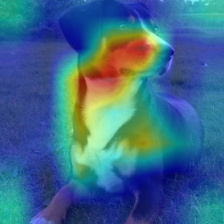} &
      \img{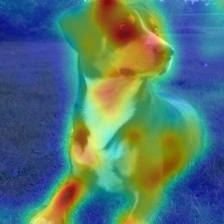} &

      \img{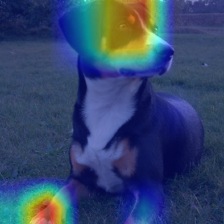} &
      \img{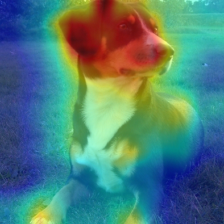} &
      \img{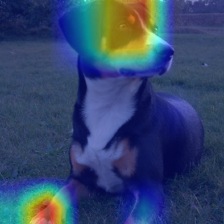} &
      \img{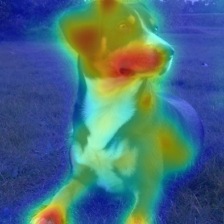}\\
      \end{tabular}
  \Caption[fig:activations-om-img-diff-supply]{ImageNet Visualization}{
    The figure shows the saliency maps generated by (from left to right) Grad-CAM, FullGradCAM, HiResCAM and FAME using different models on four ImageNet samples, evaluated with three different pre-trained networks.

  }

\end{figure*}

%% file: figures/ARface_sim_3_networks.tex
\begin{figure*}[p]
  \newcommand\img[1]{\includegraphics[width=0.06\textwidth]{#1}}
  \centering\footnotesize
    \begin{tabular}{@{}c@{}c@{\ }c@{}c@{\ }c@{}c@{\,}c@{}c@{\ }c@{}c@{\,}c@{}c@{\ }c@{}c@{\,}c@{}c@{}}
      \multicolumn{2}{c}{Grad-CAM} &
      \multicolumn{2}{@{}c@{}}{Grad-CAM-EW} &
      \multicolumn{2}{c}{CorrRISE $e_+$} &
      \multicolumn{2}{c}{CorrRISE $e_-$} &
      \multicolumn{2}{c}{FGGB $e_+$} &
      \multicolumn{2}{c}{FGGB $e_-$} &
      \multicolumn{2}{c}{FAME $e_+$} &
      \multicolumn{2}{c}{FAME $e_-$}\\

      \img{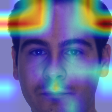} &
      \img{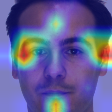} &
      \img{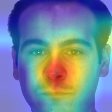} &
      \img{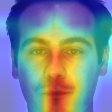}&
      \img{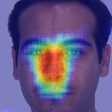}&
      \img{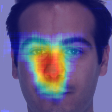}&
      \img{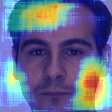}&
      \img{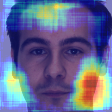}&
      \img{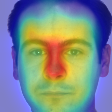}&
      \img{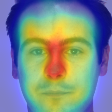}&
      \img{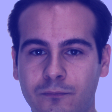}&
      \img{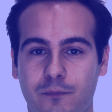}&
      \img{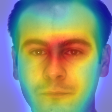}&
      \img{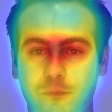}&
      \img{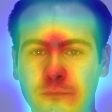}&
      \img{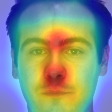}\\[-.5ex]

      \img{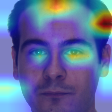} &
      \img{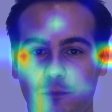} &
      \img{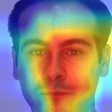} &
      \img{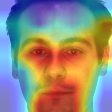}&
      \img{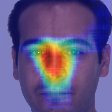}&
      \img{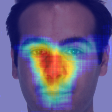}&
      \img{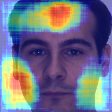}&
      \img{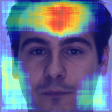}&
      \img{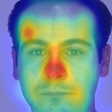}&
      \img{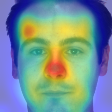}&
      \img{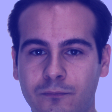}&
      \img{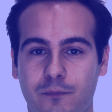}&
      \img{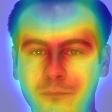}&
      \img{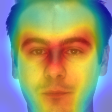}&
      \img{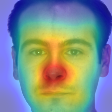}&
      \img{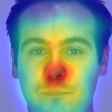}\\[-.5ex]

      \img{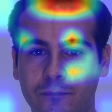} &
      \img{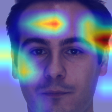} &
      \img{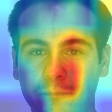} &
      \img{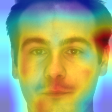}&
      \img{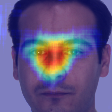}&
      \img{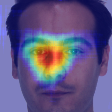}&
      \img{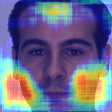}&
      \img{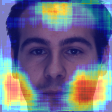}&
      \img{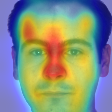}&
      \img{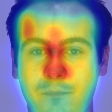}&
      \img{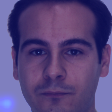}&
      \img{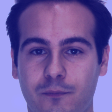}&
      \img{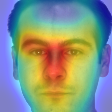}&
      \img{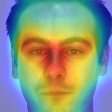}&
      \img{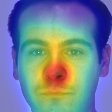}&
      \img{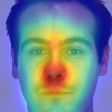}\\[.5ex]

      \multicolumn{16}{c}{(a) Protocol \texttt{neutral}}\\[1.5ex]

      \img{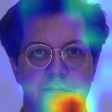} &
      \img{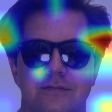} &
      \img{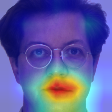} &
      \img{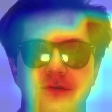}&
      \img{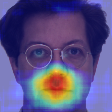}&
      \img{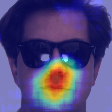}&
      \img{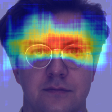}&
      \img{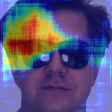}&
      \img{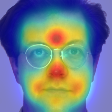}&
      \img{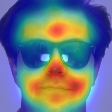}&
      \img{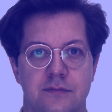}&
      \img{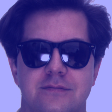}&
      \img{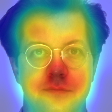}&
      \img{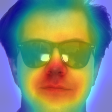}&
      \img{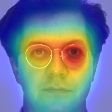}&
      \img{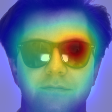}\\[-.5ex]

      \img{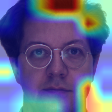} &
      \img{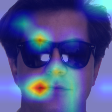} &
      \img{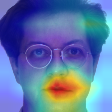} &
      \img{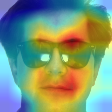}&
      \img{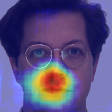}&
      \img{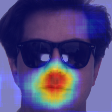}&
      \img{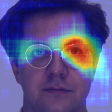}&
      \img{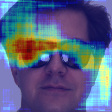}&
      \img{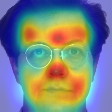}&
      \img{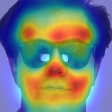}&
      \img{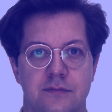}&
      \img{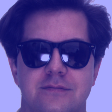}&
      \img{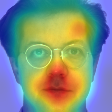}&
      \img{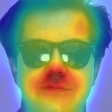}&
      \img{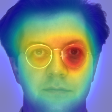}&
      \img{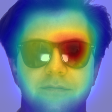}\\[-.5ex]

      \img{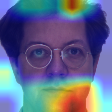} &
      \img{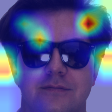} &
      \img{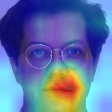} &
      \img{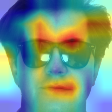}&
      \img{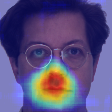}&
      \img{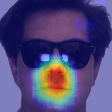}&
      \img{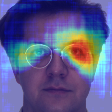}&
      \img{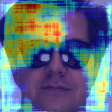}&
      \img{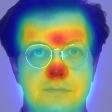}&
      \img{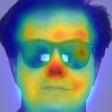}&
      \img{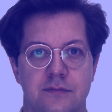}&
      \img{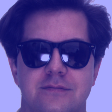}&
      \img{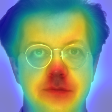}&
      \img{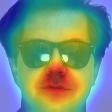}&
      \img{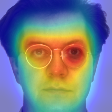}&
      \img{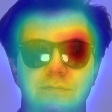}\\[.5ex]

      \multicolumn{16}{c}{(b) Protocol \texttt{glass}}\\[1.5ex]

      \img{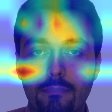} &
      \img{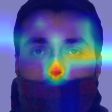} &
      \img{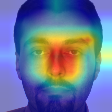} &
      \img{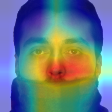}&
      \img{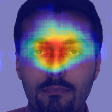}&
      \img{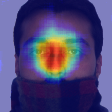}&
      \img{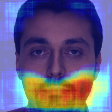}&
      \img{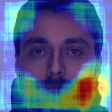}&
      \img{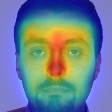}&
      \img{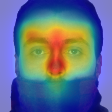}&
      \img{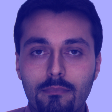}&
      \img{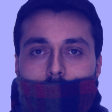}&
      \img{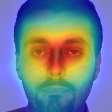}&
      \img{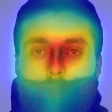}&
      \img{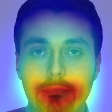}&
      \img{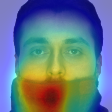}\\[-.5ex]

      \img{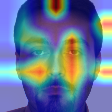} &
      \img{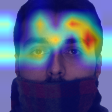} &
      \img{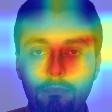} &
      \img{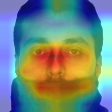}&
      \img{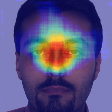}&
      \img{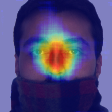}&
      \img{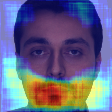}&
      \img{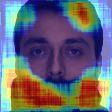}&
      \img{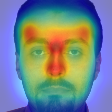}&
      \img{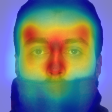}&
      \img{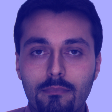}&
      \img{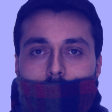}&
      \img{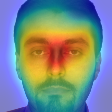}&
      \img{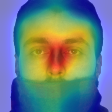}&
      \img{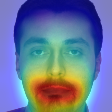}&
      \img{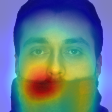}\\[-.5ex]

      \img{images/ARface/sim/Adaface_ir_101_GradCAM_scarf_Pos_6_g} &
      \img{images/ARface/sim/Adaface_ir_101_GradCAM_scarf_Pos_6_p} &
      \img{images/ARface/sim/Adaface_ir_101_GradCAMElementWise_scarf_Pos_6_g} &
      \img{images/ARface/sim/Adaface_ir_101_GradCAMElementWise_scarf_Pos_6_p}&
      \img{images/ARface/sim/Adaface_ir_101_CorrRISE_scarf_Pos_6_g_sim}&
      \img{images/ARface/sim/Adaface_ir_101_CorrRISE_scarf_Pos_6_p_sim}&
      \img{images/ARface/sim/Adaface_ir_101_CorrRISE_scarf_Pos_6_g_dissim}&
      \img{images/ARface/sim/Adaface_ir_101_CorrRISE_scarf_Pos_6_p_dissim}&
      \img{images/ARface/sim/Adaface_ir_101_fggb_scarf_Pos_6_g_sim}&
      \img{images/ARface/sim/Adaface_ir_101_fggb_scarf_Pos_6_p_sim}&
      \img{images/ARface/sim/Adaface_ir_101_fggb_scarf_Pos_6_g_dissim}&
      \img{images/ARface/sim/Adaface_ir_101_fggb_scarf_Pos_6_p_dissim}&
      \img{images/ARface/sim/Adaface_ir_101_fame_cos_scarf_Pos_6_g}&
      \img{images/ARface/sim/Adaface_ir_101_fame_cos_scarf_Pos_6_p}&
      \img{images/ARface/sim/Adaface_ir_101_fame_1-cos_scarf_Pos_6_g}&
      \img{images/ARface/sim/Adaface_ir_101_fame_1-cos_scarf_Pos_6_p}\\[.5ex]
      \multicolumn{16}{c}{(c) Protocol \texttt{scarf}}
      \end{tabular}

  \Caption[fig:activations-fr-arface]{ARface}{
    The figure shows a comparative visualization of explanation maps generated on AR face for genuine (same-identity) image pairs using three networks IResNet18, IResNet50 and IResNet101.
    XAI techniques include Grad-CAM, Grad-CAM-EW, CorrRISE, FGGB, and FAME, including similar $e_{+}$ and dissimilar attribution $e_{-}$ where appropriate.

  }

\end{figure*}

%% file: figures/SCface_sim_3_networks.tex
\begin{figure*}[p]
  \newcommand\img[1]{\includegraphics[width=0.06\textwidth]{#1}}
  \centering\footnotesize
    \begin{tabular}{@{}c@{}c@{\ }c@{}c@{\ }c@{}c@{\,}c@{}c@{\ }c@{}c@{\,}c@{}c@{\ }c@{}c@{\,}c@{}c@{}}
      \multicolumn{2}{c}{Grad-CAM} &
      \multicolumn{2}{@{}c@{}}{Grad-CAM-EW} &
      \multicolumn{2}{c}{CorrRISE $e_+$} &
      \multicolumn{2}{c}{CorrRISE $e_-$} &
      \multicolumn{2}{c}{FGGB $e_+$} &
      \multicolumn{2}{c}{FGGB $e_-$} &
      \multicolumn{2}{c}{FAME $e_+$} &
      \multicolumn{2}{c}{FAME $e_-$}\\

      \img{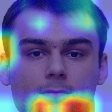} &
      \img{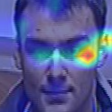} &
      \img{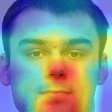} &
      \img{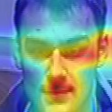}&
      \img{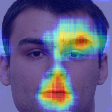}&
      \img{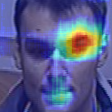}&
      \img{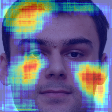}&
      \img{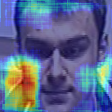}&
      \img{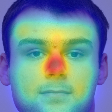}&
      \img{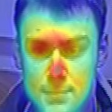}&
      \img{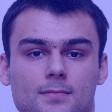}&
      \img{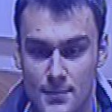}&
      \img{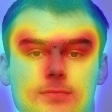}&
      \img{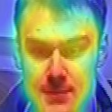}&
      \img{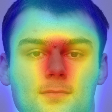}&
      \img{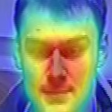}\\[-.5ex]

      \img{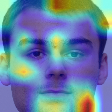} &
      \img{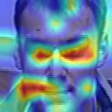} &
      \img{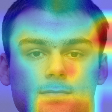} &
      \img{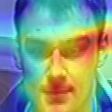}&
      \img{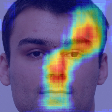}&
      \img{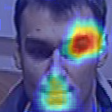}&
      \img{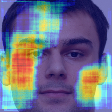}&
      \img{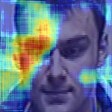}&
      \img{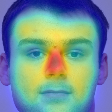}&
      \img{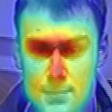}&
      \img{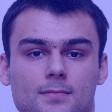}&
      \img{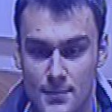}&
      \img{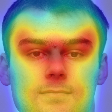}&
      \img{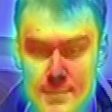}&
      \img{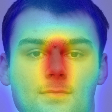}&
      \img{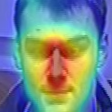}\\[-.5ex]

      \img{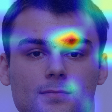} &
      \img{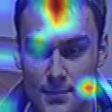} &
      \img{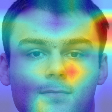} &
      \img{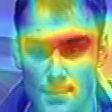}&
      \img{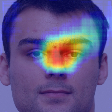}&
      \img{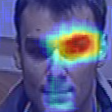}&
      \img{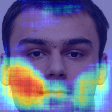}&
      \img{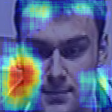}&
      \img{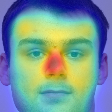}&
      \img{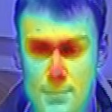}&
      \img{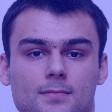}&
      \img{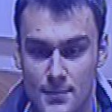}&
      \img{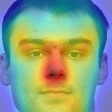}&
      \img{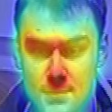}&
      \img{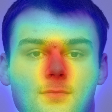}&
      \img{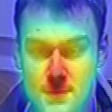}\\[.5ex]

      \multicolumn{16}{c}{(a) Protocol \texttt{close}}\\[1.5ex]

      \img{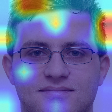} &
      \img{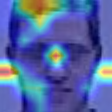} &
      \img{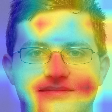} &
      \img{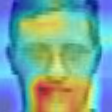}&
      \img{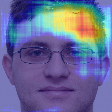}&
      \img{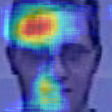}&
      \img{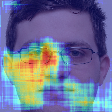}&
      \img{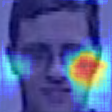}&
      \img{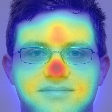}&
      \img{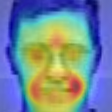}&
      \img{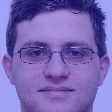}&
      \img{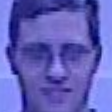}&
      \img{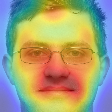}&
      \img{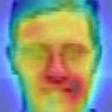}&
      \img{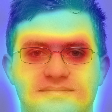}&
      \img{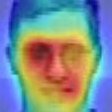}\\[-.5ex]

      \img{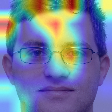} &
      \img{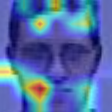} &
      \img{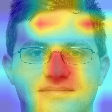} &
      \img{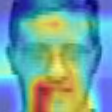}&
      \img{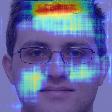}&
      \img{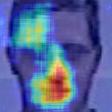}&
      \img{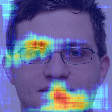}&
      \img{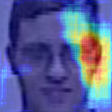}&
      \img{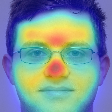}&
      \img{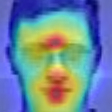}&
      \img{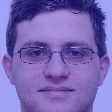}&
      \img{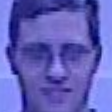}&
      \img{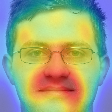}&
      \img{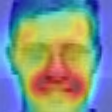}&
      \img{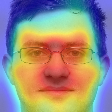}&
      \img{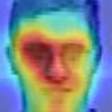}\\[-.5ex]

      \img{images/SCface/sim/Adaface_ir_101_GradCAM_medium_Pos_17_g} &
      \img{images/SCface/sim/Adaface_ir_101_GradCAM_medium_Pos_17_p} &
      \img{images/SCface/sim/Adaface_ir_101_GradCAMElementWise_medium_Pos_17_g} &
      \img{images/SCface/sim/Adaface_ir_101_GradCAMElementWise_medium_Pos_17_p}&
      \img{images/SCface/sim/Adaface_ir_101_CorrRISE_medium_Pos_17_g_sim}&
      \img{images/SCface/sim/Adaface_ir_101_CorrRISE_medium_Pos_17_p_sim}&
      \img{images/SCface/sim/Adaface_ir_101_CorrRISE_medium_Pos_17_g_dissim}&
      \img{images/SCface/sim/Adaface_ir_101_CorrRISE_medium_Pos_17_p_dissim}&
      \img{images/SCface/sim/Adaface_ir_101_fggb_medium_Pos_17_g_sim}&
      \img{images/SCface/sim/Adaface_ir_101_fggb_medium_Pos_17_p_sim}&
      \img{images/SCface/sim/Adaface_ir_101_fggb_medium_Pos_17_g_dissim}&
      \img{images/SCface/sim/Adaface_ir_101_fggb_medium_Pos_17_p_dissim}&
      \img{images/SCface/sim/Adaface_ir_101_fame_cos_medium_Pos_17_g}&
      \img{images/SCface/sim/Adaface_ir_101_fame_cos_medium_Pos_17_p}&
      \img{images/SCface/sim/Adaface_ir_101_fame_1-cos_medium_Pos_17_g}&
      \img{images/SCface/sim/Adaface_ir_101_fame_1-cos_medium_Pos_17_p}\\[.5ex]

      \multicolumn{16}{c}{(b) Protocol \texttt{medium}}\\[1.5ex]

      \img{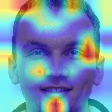} &
      \img{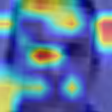} &
      \img{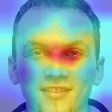} &
      \img{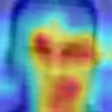}&
      \img{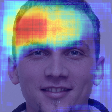}&
      \img{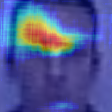}&
      \img{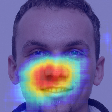}&
      \img{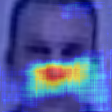}&
      \img{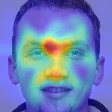}&
      \img{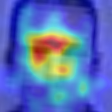}&
      \img{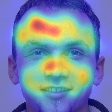}&
      \img{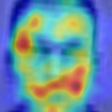}&
      \img{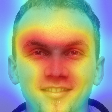}&
      \img{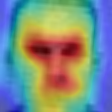}&
      \img{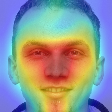}&
      \img{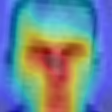}\\[-.5ex]

      \img{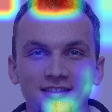} &
      \img{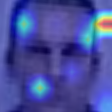} &
      \img{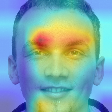} &
      \img{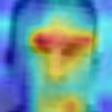}&
      \img{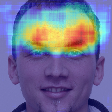}&
      \img{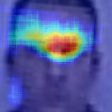}&
      \img{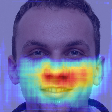}&
      \img{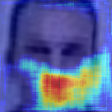}&
      \img{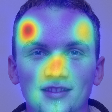}&
      \img{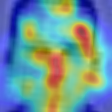}&
      \img{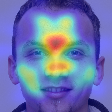}&
      \img{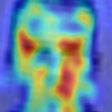}&
      \img{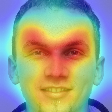}&
      \img{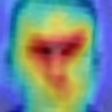}&
      \img{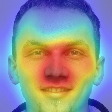}&
      \img{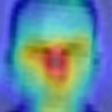}\\[-.5ex]

      \img{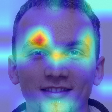} &
      \img{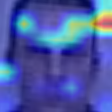} &
      \img{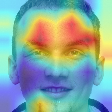} &
      \img{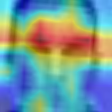}&
      \img{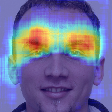}&
      \img{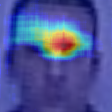}&
      \img{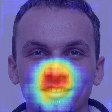}&
      \img{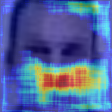}&
      \img{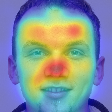}&
      \img{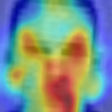}&
      \img{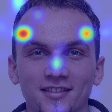}&
      \img{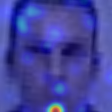}&
      \img{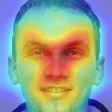}&
      \img{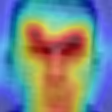}&
      \img{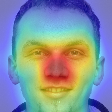}&
      \img{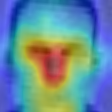}\\[.5ex]

      \multicolumn{16}{c}{(c) Protocol \texttt{far}}
    \end{tabular}

  \Caption[fig:activations-fr-scface]{SCface}{
    The figure shows a comparative visualization of explanation maps generated on SCface for genuine (same-identity) image pairs using three networks IResNet18, IResNet50 and IResNet101.
    XAI techniques include Grad-CAM, Grad-CAM-EW, CorrRISE, FGGB, and FAME, including similar $e_{+}$ and dissimilar attribution $e_{-}$ where appropriate.
  }

\end{figure*}

%% file: figures/CFP_sim_3_networks.tex
\begin{figure*}[p]
  \newcommand\img[1]{\includegraphics[width=0.06\textwidth]{#1}}
  \centering\footnotesize
    \begin{tabular}{@{}c@{}c@{\ }c@{}c@{\ }c@{}c@{\,}c@{}c@{\ }c@{}c@{\,}c@{}c@{\ }c@{}c@{\,}c@{}c@{}}
      \multicolumn{2}{c}{Grad-CAM} &
      \multicolumn{2}{c}{Grad-CAM-EW} &
      \multicolumn{2}{c}{CorrRISE $e_+$} &
      \multicolumn{2}{c}{CorrRISE $e_-$} &
      \multicolumn{2}{c}{FGGB $e_+$} &
      \multicolumn{2}{c}{FGGB $e_-$} &
      \multicolumn{2}{c}{FAME $e_+$} &
      \multicolumn{2}{c}{FAME $e_-$}\\

      \img{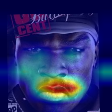} &
      \img{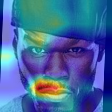} &
      \img{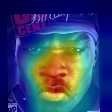} &
      \img{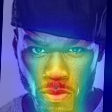}&
      \img{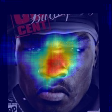}&
      \img{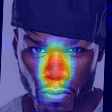}&
      \img{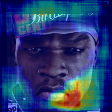}&
      \img{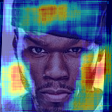}&
      \img{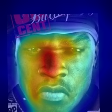}&
      \img{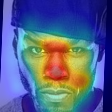}&
      \img{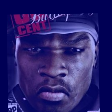}&
      \img{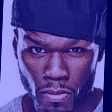}&
      \img{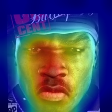}&
      \img{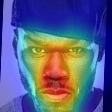}&
      \img{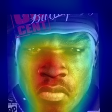}&
      \img{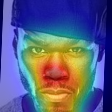}\\[-.5ex]

      \img{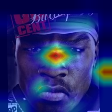} &
      \img{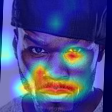} &
      \img{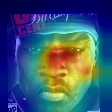} &
      \img{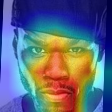}&
      \img{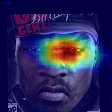}&
      \img{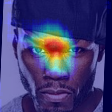}&
      \img{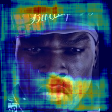}&
      \img{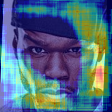}&
      \img{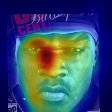}&
      \img{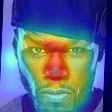}&
      \img{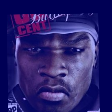}&
      \img{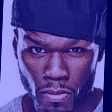}&
      \img{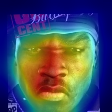}&
      \img{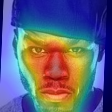}&
      \img{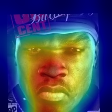}&
      \img{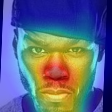}\\[-.5ex]

      \img{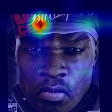} &
      \img{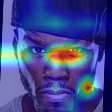} &
      \img{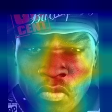} &
      \img{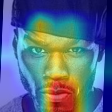}&
      \img{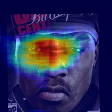}&
      \img{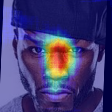}&
      \img{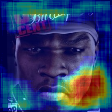}&
      \img{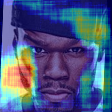}&
      \img{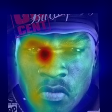}&
      \img{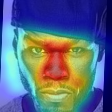}&
      \img{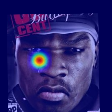}&
      \img{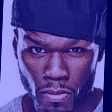}&
      \img{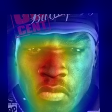}&
      \img{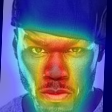}&
      \img{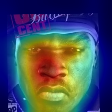}&
      \img{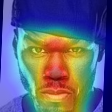}\\[.5ex]

      \multicolumn{16}{c}{(a) Protocol \texttt{FF}}\\[1.5ex]

      \img{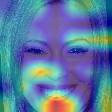} &
      \img{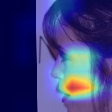} &
      \img{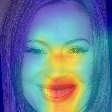} &
      \img{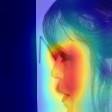}&
      \img{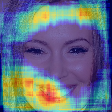}&
      \img{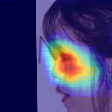}&
      \img{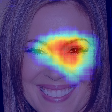}&
      \img{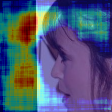}&
      \img{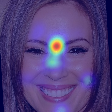}&
      \img{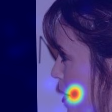}&
      \img{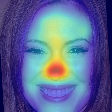}&
      \img{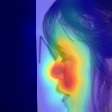}&
      \img{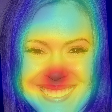}&
      \img{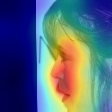}&
      \img{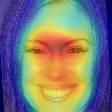}&
      \img{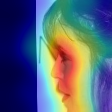}\\[-.5ex]

      \img{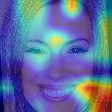} &
      \img{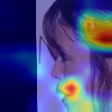} &
      \img{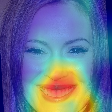} &
      \img{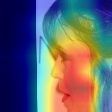}&
      \img{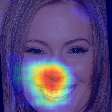}&
      \img{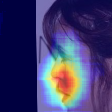}&
      \img{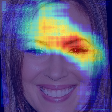}&
      \img{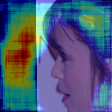}&
      \img{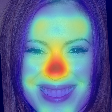}&
      \img{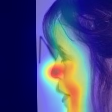}&
      \img{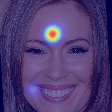}&
      \img{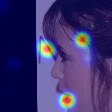}&
      \img{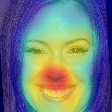}&
      \img{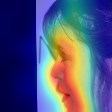}&
      \img{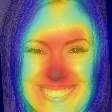}&
      \img{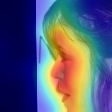}\\[-.5ex]

      \img{images/CFP/sim/Adaface_ir_101_GradCAM_01FP_Pos_505_g} &
      \img{images/CFP/sim/Adaface_ir_101_GradCAM_01FP_Pos_505_p} &
      \img{images/CFP/sim/Adaface_ir_101_GradCAMElementWise_01FP_Pos_505_g} &
      \img{images/CFP/sim/Adaface_ir_101_GradCAMElementWise_01FP_Pos_505_p}&
      \img{images/CFP/sim/Adaface_ir_101_CorrRISE_01FP_Pos_505_g_sim}&
      \img{images/CFP/sim/Adaface_ir_101_CorrRISE_01FP_Pos_505_p_sim}&
      \img{images/CFP/sim/Adaface_ir_101_CorrRISE_01FP_Pos_505_g_dissim}&
      \img{images/CFP/sim/Adaface_ir_101_CorrRISE_01FP_Pos_505_p_dissim}&
      \img{images/CFP/sim/Adaface_ir_101_fggb_01FP_Pos_505_g_sim}&
      \img{images/CFP/sim/Adaface_ir_101_fggb_01FP_Pos_505_p_sim}&
      \img{images/CFP/sim/Adaface_ir_101_fggb_01FP_Pos_505_g_dissim}&
      \img{images/CFP/sim/Adaface_ir_101_fggb_01FP_Pos_505_p_dissim}&
      \img{images/CFP/sim/Adaface_ir_101_fame_cos_01FP_Pos_505_g}&
      \img{images/CFP/sim/Adaface_ir_101_fame_cos_01FP_Pos_505_p}&
      \img{images/CFP/sim/Adaface_ir_101_fame_1-cos_01FP_Pos_505_g}&
      \img{images/CFP/sim/Adaface_ir_101_fame_1-cos_01FP_Pos_505_p}\\[.5ex]

      \multicolumn{16}{c}{(b) Protocol \texttt{FP}}

    \end{tabular}

  \Caption[fig:activations-fr-cfp]{CFP}{
    The figure shows a comparative visualization of explanation maps generated on CFP dataset for genuine (same-identity) image pairs using three networks IResNet18, IResNet50 and IResNet101.
    XAI techniques include Grad-CAM, Grad-CAM-EW, CorrRISE, FGGB, and FAME, including similar $e_{+}$ and dissimilar attribution $e_{-}$ where appropriate.

  }

\end{figure*}

%% file: sections/Publications.bib
@inproceedings{rozsa2017lots,
  title={{LOTS} about attacking deep features},
  author={Rozsa, Andras and G\"unther, Manuel and Boult, Terranee E.},
  booktitle={International Joint Conference on Biometrics (IJCB)},
  optpages={168--176},
  year={2017},
  organization={IEEE}
}

@article{pereira20228years,
  title   = {Eight Years of Face Recognition Research: Reproducibility, Achievements and Open Issues},
  author  = {de Freitas Pereira, Tiago and Schmidli, Dominic and Linghu, Yu and Zhang, Xinyi and Marcel, S\'ebastien and G\"unther, Manuel},
  year    = {2022},
  journal = {arXiv}
}


%% file: sections/References.bib
@inproceedings{krizhevsky2012alexnet,
  title = {{ImageNet} Classification with Deep Convolutional Neural Networks},
  author = {Krizhevsky, Alex  and Sutskever, Ilya and Hinton, Geoffrey E.},
  booktitle = {Advances in Neural Information Processing Systems (NeurIPS)},
  year = {2012},
}

@InProceedings{goodfellow2015explaining,
  author={Goodfellow, Ian J. and Shlens, Jonathon and Szegedy, Christian},
  title={Explaining and Harnessing Adversarial Examples},
  booktitle={International Conference on Learning Representation (ICLR)},
  year={2015},
}

@inproceedings{deng2009imagenet,
  title={{ImageNet}: A large-scale hierarchical image database},
  author={Deng, Jia and Dong, Wei and Socher, Richard and Li, Li-Jia and Li, Kai and Fei-Fei, Li},
  booktitle={Conference on Computer Vision and Pattern Recognition (CVPR)},
  year={2009},
  organization={IEEE}
}

@inbook{phillips2011evaluation,
  author={Phillips, P. Jonathon and Grother, Patrick and Micheals, Ross},
  chapter={Evaluation Methods in Face Recognition},
  title={Handbook of Face Recognition},
  year={2011},
  publisher={Springer},
  editor={Li, Stan Z. and Jain, Anil K.},
  edition={2nd}
}

@inproceedings{he2016deep,
  title={Deep Residual Learning for Image Recognition},
  author={He, Kaiming and Zhang, Xiangyu and Ren, Shaoqing and Sun, Jian},
  booktitle={Conference on Computer Vision and Pattern Recognition (CVPR)},
  year={2016},
  organization={IEEE}
}

@inproceedings{zeiler2014visualizing,
  title={Visualizing and understanding convolutional networks},
  author={Zeiler, Matthew D. and Fergus, Rob},
  booktitle={European Conference on Computer Vision (ECCV)},
  optpages={818--833},
  year={2014},
  optorganization={Springer}
}

@inproceedings{deng2019arcface,
  title={Arcface: Additive angular margin loss for deep face recognition},
  author={Deng, Jiankang and Guo, Jia and Xue, Niannan and Zafeiriou, Stefanos},
  booktitle={Conference on Computer Vision and Pattern Recognition (CVPR)},
  year={2019}
}

@article{grgic2011scface,
  author = {Grgic, Mislav and Delac, Kresimir and Grgic, Sonja},
  title = {{SCface} - surveillance cameras face database},
  journal = {Multimedia Tools and Applications},
  number = 3,
  volume = 51,
  year = 2011
}

@inproceedings{pytorch,
  title = {{PyTorch}: An Imperative Style, High-Performance Deep Learning Library},
  author = {Paszke, Adam and Gross, Sam and Massa, Francisco and Lerer, Adam and Bradbury, James and Chanan, Gregory and Killeen, Trevor and Lin, Zeming and Gimelshein, Natalia and Antiga, Luca and Desmaison, Alban and Kopf, Andreas and Yang, Edward and DeVito, Zachary and Raison, Martin and Tejani, Alykhan and Chilamkurthy, Sasank and Steiner, Benoit and Fang, Lu and Bai, Junjie and Chintala, Soumith},
  booktitle = {Advances in Neural Information Processing Systems (NeuRIPS)},
  year = {2019},
}

@article{wallace2012cross,
  author   = {Wallace, Roy and McLaren, Mitchell and McCool, Chris and Marcel, S\'ebastien},
  title    = {Cross-pollination of normalisation techniques from speaker to face authentication using {Gaussian} mixture models},
  journal  = {Transactions on Information Forensics and Security (TIFS)},
  volume   = {7},
  number   = {2},
  year     = {2012}
}

@inproceedings{sengupta2016cfp,
  author    = {Sengupta, Soumyadip and Chen, Jun-Cheng and Castillo, Carlos and Patel, Vishal M and Chellappa, Rama and Jacobs, David W},
  booktitle = {Winter Conference on Applications of Computer Vision (WACV)},
  title     = {Frontal to Profile Face Verification in the Wild},
  year      = {2016}
}

@inproceedings{meng2021magface,
  title     = {{MagFace}: A universal representation for face recognition and quality assessment},
  author    = {Meng, Qiang and Zhao, Shichao and Huang, Zhida and Zhou, Feng},
  booktitle = {Conference on Computer Vision and Pattern Recognition (CVPR)},
  year      = {2021}
}

@inproceedings{selvaraju2017gradcam,
  author    = {Selvaraju, Ramprasaath R. and Cogswell, Michael and Das, Abhishek and Vedantam, Ramakrishna and Parikh, Devi and Batra, Dhruv},
  title     = {Grad-{CAM}: Visual Explanations From Deep Networks via Gradient-Based Localization},
  booktitle = {International Conference on Computer Vision (ICCV)},
  year      = {2017},
  organization = {IEEE}
}

@inproceedings{kim2022adaface,
  title     = {{AdaFace}: Quality Adaptive Margin for Face Recognition},
  author    = {Kim, Minchul and Jain, Anil K. and Liu, Xiaoming},
  booktitle = {Conference on Computer Vision and Pattern Recognition (CVPR)},
  year      = {2022}
}

@inproceedings{chattopadhay2018grad-cam++,
  author    = {Chattopadhay, Aditya and Sarkar, Anirban and Howlader, Prantik and Balasubramanian, Vineeth N},
  booktitle = {Winter Conference on Applications of Computer Vision (WACV)},
  title     = {{Grad-CAM++}: Generalized Gradient-Based Visual Explanations for Deep Convolutional Networks},
  year      = {2018},
}

@inproceedings{fong2017interpretable,
  title={Interpretable explanations of black boxes by meaningful perturbation},
  author={Fong, Ruth C and Vedaldi, Andrea},
  booktitle={International Conference on Computer Vision (ICCV)},
  year={2017}
}

@inproceedings{petsiuk2018rise,
  author    = {Petsiuk, Vitali and Das, Abir and Saenko, Kate},
  title     = {{RISE}: Randomized Input Sampling for Explanation of Black-box Models},
  booktitle = {British Machine Vision Conference (BMVC)},
  year      = {2018},
}

@techreport{martinez1998arface,
  title    = {The {AR} Face Database},
  author   = {Mart\'inez, Aleix M. and Benavente, Robert},
  year     = {1998},
  institution = {Universitat Aut\`onoma de Barcelona, {CVC}},
  number = {24}
}

@inproceedings{rong2022road,
  title     = {A Consistent and Efficient Evaluation Strategy for Attribution Methods},
  author    = {Rong, Yao and Leemann, Tobias and Borisov, Vadim and Kasneci, Gjergji and Kasneci, Enkelejda},
  booktitle = {International Conference on Machine Learning (ICML)},
  year      = {2022},
  publisher = {PMLR}
}

@article{tucci2024overview,
  author = {Tucci, Cesare and Della Greca, Attilio and Tortora, Genoveffa and Francese, Rita},
  title = {Explainable biometrics: A systematic literature review},
  journal = {Journal of Ambient Intelligence and Humanized Computing},
  year = {2024}
}

@article{draelos2020hirescam,
  title   = {Use {HiResCAM} instead of {Grad-CAM} for faithful explanations of convolutional neural networks},
  author  = {Draelos, Rachel Lea and Carin, Lawrence},
  journal = {arXiv},
  year    = {2020}
}

@inproceedings{srinivas2019fullgrad,
    title={Full-Gradient Representation for Neural Network Visualization},
    author={Srinivas, Suraj and Fleuret, François},
    booktitle={Advances in Neural Information Processing Systems (NeurIPS)},
    year={2019}
}

@article{deng2022arcface,
  author   = {Deng, Jiankang and Guo, Jia and Yang, Jing and Xue, Niannan and Kotsia, Irene and Zafeiriou, Stefanos},
  journal  = {Transactions on Pattern Analysis and Machine Intelligence (TPAMI)},
  title    = {ArcFace: Additive Angular Margin Loss for Deep Face Recognition},
  year     = {2022},
  volume   = {44},
  number   = {10},
  pages    = {5962-5979},
  doi      = {10.1109/TPAMI.2021.3087709}
}

@article{robbins2024daliid,
  author   = {Robbins, Wes and Bertocco, Gabriel and Boult, Terrance E.},
  journal  = {IEEE Access},
  title    = {{DaliID}: Distortion-Adaptive Learned Invariance for Identification – a Robust Technique for Face Recognition and Person Re-Identification},
  year     = {2024},
  doi      = {10.1109/ACCESS.2024.3385782}
}

@inproceedings{li2025diffcam,
  author={Li, Xingjian and Zhao, Qiming and Bisht, Neelesh and Uddin, Mostofa Rafid and Yu Kim, Jin and Zhang, Bryan and Xu, Min},
  title={{DiffCAM}: Data-Driven Saliency Maps by Capturing Feature Differences},
  booktitle={Conference on Computer Vision and Pattern Recognition (CVPR)},
  year={2025},
}

@article{rudin2018stop,
  title={Stop explaining black box machine learning models for high stakes decisions and use interpretable models instead},
  author={Rudin, Cynthia},
  journal={Nature Machine Intelligence},
  year={2018},
  volume={1},
}

@article{nauta2023anecdotal,
  author = {Nauta, Meike and Trienes, Jan and Pathak, Shreyasi and Nguyen, Elisa and Peters, Michelle and Schmitt, Yasmin and Schl\"{o}tterer, J\"{o}rg and van Keulen, Maurice and Seifert, Christin},
  title = {From Anecdotal Evidence to Quantitative Evaluation Methods: A Systematic Review on Evaluating Explainable {AI}},
  year = {2023},
  journal = {ACM Computing Surveys},
}

@INPROCEEDINGS{zhou2016localization,
  author={Zhou, Bolei and Khosla, Aditya and Lapedriza, Agata and Oliva, Aude and Torralba, Antonio},
  booktitle={Conference on Computer Vision and Pattern Recognition (CVPR)},
  title={Learning Deep Features for Discriminative Localization},
  year={2016},
}

@article{longo2024xai2.0,
  title = {Explainable Artificial Intelligence {(XAI)} 2.0: A manifesto of open challenges and interdisciplinary research directions},
  journal = {Information Fusion},
  volume = {106},
  year = {2024},
  author = {Longo, Luca  and Brcic, Mario and Cabitza, Federico and Choi, Jaesik and  Confalonieri, Roberto and  Del Ser, Javier and Guidotti, Riccardo and Hayashi, Yoichi and Herrera, Francisco and Holzinger, Andreas and Jiang, Richard and Khosravi, Hassan and Lecue, Freddy and Malgieri, Gianclaudio and Páez, Andrés and Samek, Wojciech and Schneider, Johannes and Speith, Timo and Stumpf, Simone},
}

@inproceedings{lu2024corrrise,
  author    = {Lu, Yuhang and Xu, Zewei and Ebrahimi, Touradj},
  title     = {Towards Visual Saliency Explanations of Face Verification},
  booktitle = {Winter Conference on Applications of Computer Vision (WACV)},
  year      = {2024},
}

@inproceedings{adebayo2018sanity,
  author = {Adebayo, Julius and Gilmer, Justin and Muelly, Michael and Goodfellow, Ian and Hardt, Moritz and Kim, Been},
  title = {Sanity checks for saliency maps},
  year = {2018},
  booktitle = {Advances in Neural Information Processing Systems (NeurIPS)},
}

@article{ivanovs2021perturbation,
  title    = {Perturbation-based methods for explaining deep neural networks: A survey},
  journal  = {Pattern Recognition Letters},
  volume   = {150},
  pages    = {228-234},
  year     = {2021},
  issn     = {0167-8655},
  doi   = {https://doi.org/10.1016/j.patrec.2021.06.030},
  url   = {https://www.sciencedirect.com/science/article/pii/S0167865521002440},
  author   = {Ivanovs, Maksims and Kadikis, Roberts and Ozols, Kaspars}
}

@article{zhu2021visual,
  title={Visual explanation for deep metric learning},
  author={Zhu, Sijie and Yang, Taojiannan and Chen, Chen},
  journal={Transactions on Image Processing (TIP)},
  volume={30},
  pages={7593--7607},
  year={2021},
  publisher={IEEE}
}

@inproceedings{lu2024fggb,
  title        = {Explainable Face Verification via Feature-Guided Gradient Backpropagation},
  author       = {Lu, Yuhang and Xu, Zewei and Ebrahimi, Touradj},
  booktitle    = {International Conference on Automatic Face and Gesture Recognition (FG)},
  year         = {2024},
  organization = {IEEE}
}

@inproceedings{huber2024xssab,
  title     = {Efficient Explainable Face Verification based on Similarity Score Argument Backpropagation},
  author    = {Huber, Marco and Luu, Anh Thi and Terh{\"o}rst, Philipp and Damer, Naser},
  booktitle = {Winter Conference on Applications of Computer Vision (WACV)},
  year      = {2024}
}
